\pdfoutput=1
\documentclass[5p,times,twocolumn]{elsarticle}

\usepackage[T1]{fontenc}
\usepackage{amsmath}
\usepackage{amssymb}
\usepackage{graphicx}
\usepackage{subcaption}
\usepackage{tikz}
\usepackage{xcolor}
\usepackage{url}
\usepackage[colorlinks=true,linkcolor=blue!60!black,citecolor=blue!60!black,urlcolor=blue!60!black]{hyperref}

\def\checkmark{\tikz\fill[scale=0.4](0,.35) -- (.25,0) -- (1,.7) -- (.25,.15) -- cycle;}

\newcommand{\GG}{\mathcal{G}}
\newcommand{\KK}{\mathcal{K}}
\newcommand{\CP}{\Pi^C_{\theta}}
\newcommand{\MP}{\Pi^M_{\theta}}
\newcommand{\AP}{\Pi^A_{\theta}}
\newcommand{\axclass}{\phi_{\text{class}}}
\newcommand{\axmacro}{\phi_{\text{macro}}}
\newcommand{\axmasked}{\phi_{\text{masked}}}
\newcommand{\axattr}{\phi_{\text{attr}}}
\newcommand{\axsimil}{\phi_{\text{simil}}}
\newcommand{\axdissimil}{\phi_{\text{dissimil}}}

\journal{Pattern Recognition}

\begin{document}

\begin{frontmatter}

\title{Boosting zero-shot learning through neuro-symbolic integration}

\author[1]{Francesco {Manigrasso}}
\ead{francesco.manigrasso@polito.it}
\author[1]{Fabrizio {Lamberti}}
\ead{fabrizio.lamberti@polito.it}
\author[1]{Lia {Morra}}
\ead{lia.morra@polito.it}

\address[1]{Politecnico di Torino, Corso Duca degli Abruzzi 24, Turin, 10129, Italy}

\begin{abstract}
Zero-shot learning (ZSL) aims to train deep neural networks to recognize objects from unseen classes, starting from a semantic description of the concepts. Neuro-symbolic (NeSy) integration refers to a class of techniques that incorporate symbolic knowledge representation and reasoning with the learning capabilities of deep neural networks. However, to date, few studies have explored how to leverage NeSy techniques to inject prior knowledge during the training process to boost ZSL capabilities. Here, we present Fuzzy Logic Prototypical Network (FLPN) that formulates the classification task as prototype matching in a visual-semantic embedding space, which is trained by optimizing a NeSy loss. Specifically, FLPN exploits the Logic Tensor Network (LTN) framework to incorporate background knowledge in the form of logical axioms by grounding a first-order logic language as differentiable operations between real tensors. This prior knowledge includes class hierarchies (classes and macroclasses) along with robust high-level inductive biases. The latter allow, for instance, to handle exceptions in class-level attributes and to enforce similarity between images of the same class, preventing premature overfitting to seen classes and improving overall performance.
FLPN learns both class-level and attribute-level prototypes, the latter through an attention mechanism specialized for either convolutional- or transformer-based backbones. FLPN achieves state-of-the-art performance on the GZSL benchmarks AWA2 and SUN, matching or exceeding the performance of competing algorithms with minimal computational overhead.
The code is available at \url{https://github.com/FrancescoManigrass/FLPN}.
\end{abstract}


\end{frontmatter}

\section{Introduction}
\label{sec:intro}

The field of computer vision stands to benefit from methods that promote the combination of learning and reasoning \citep{yu2021survey,LTN,van2021modular}.  Neuro-symbolic (NeSy) artificial intelligence aims to integrate symbolic representation and reasoning with perception and machine learning within a unified framework, enhancing the explainability, robustness, and flexibility of deep learning models. We focus here in particular on a class of NeSy techniques that aim to incorporate previous knowledge expressed in symbolic form into the training objectives of deep neural networks, compensating for imperfect or limited supervision provided by the labeled examples \citep{manigrasso2021faster,donadello2019compensating,Badreddine2023logLTNDF, Sikka2020ZeroShotLW, Martone2022PROTOtypicalLT,Manigrasso2023FuzzyLV}. This is particularly relevant in Zero-Shot Learning (ZSL), which aims to recognize novel objects by linking both known and unknown categories with some kind of external information, usually in the form of class attributes or descriptions \citep{awa2}. In this paper, we introduce the Fuzzy Logic Prototypical Network (FLPN), the latest addition to a series of NeSy approaches for ZSL image recognition that includes Proto-LTN \citep{Martone2022PROTOtypicalLT} and FLVN \citep{Manigrasso2023FuzzyLV}.

The effectiveness of neural networks stems from their ability to learn distributed representations that, unlike symbolic representations, induce a rich similarity space, in which semantically close concepts (or inputs) are close in distance \cite{goodfellow2016deep}. FLPN builds on Logic Tensor Networks (LTNs) \citep{LTN}, which can be classified as tightly-coupled, distributed NeSy architectures (classified as Level 5 according to Kautz's taxonomy \cite{garcez2023neurosymbolic}). In FLPN, the NeSy component acts as a soft constraint or regularizer, guiding the network towards learning a more effective representation, which is exploited at inference time to classify new samples.  FLPN thus seeks to surpass the dichotomy between symbolic and distributed representations by associating symbols (e.g., class labels) with prototypes (a process denoted as \textit{grounding}). This approach offers several advantages: the ability to exploit semantic relationships between symbols through distance, easier manipulation of abstract concepts in vectorized form, and improved interpretability of prototypes over simple labels through visualization techniques like t-SNE \cite{Martone2022PROTOtypicalLT}.

Briefly, the core idea of LTN is that each logical axiom (e.g., class hierarchies, attribute constraints) becomes a differentiable constraint incorporated into the loss function. This ensures that the neural network not only learns from the supervised signal (i.e., the seen classes), but also must satisfy the symbolic constraints, which enforce a proper organization of the prototype and feature space. Specifically, in FLPN, class hierarchies (macro-classes) provide a coherent semantic structure, guiding the network to ``place'' unseen classes appropriately when they share characteristics with seen classes. The attribute axioms and the masking mechanism prevent the network from over-relying or misrepresenting  attributes that do not appear in all examples of a class, thereby improving the separability of different attributes in the embedding space. Similarity axioms promote consistent clustering of images within the feature space. Overall, these symbolic constraints act as an inductive bias, restricting the solution space and reducing the risk of overfitting to the training (seen) classes.

In summary, our contributions are the following ones:
\begin{itemize}
    \item we introduce the FLPN architecture, a novel NeSy approach to ZSL recognition that combines the flexibility of prototypical networks with a semantics-enhanced training procedure empowered by the LTN framework. FLPN  combines elements from our previous works \citep{Martone2022PROTOtypicalLT,Manigrasso2023FuzzyLV} and extends them with the extraction of attribute-level information and a more comprehensive set of axioms;
    \item through experiments on three different benchmarks, we show that FLPN outperforms or closely matches existing ZSL techniques with minimal overhead at inference time and without the need to generate additional samples at training time;
    \item we conduct extensive experiments and ablation studies to understand the role of different backbones (CNNs and transformers), LTN semantics, and  knowledge base formulations.
\end{itemize}

The rest of the paper is organized as follows. Section~\ref{sec:releated} presents the foundational concepts of the LTN framework and related studies. Section~\ref{sec:flpn} describes the proposed architecture. The performance of the model in the ZSL and GZSL contexts over established benchmarks is evaluated in Section~\ref{sec:experiments} and Section~\ref{sec:results}. Finally, Section~\ref{sec:conclusion} discusses the results and potential future research directions.

\section{Related work}
\label{sec:releated}

\subsection{Zero-Shot Learning}
ZSL is a machine learning paradigm in which a model is trained to recognize and classify objects from unseen categories by leveraging auxiliary information, such as semantic attributes, that relates unseen categories to seen training categories. Generalized Zero-Shot Learning (GZSL) extends the ZSL paradigm to recognize and classify objects from seen and unseen categories. Unlike ZSL, GZSL requires the model to effectively handle a more complicated scenario by having to distinguish seen and unseen classes in the same space. Several methods have been developed for ZSL, which we will roughly categorize in the following as embedding-based (either with fixed embeddings or trained end-to-end) and generative methods.

In \textbf{embedding-based} methods, semantic attributes and visual data are integrated by mapping them into a shared space. Some techniques project images into the attribute space through an embedding function, treating semantic attributes as the common space~\citep{CC-ZSL,zhou2023attribute}. FLPN projects instead the semantic attributes into the image embedding space using a projection function and uses the image embedding space as the common reference space, similarly to many methods dating back to DEM, in order to address the \textit{hubness} problem~\citep{Martone2022PROTOtypicalLT,dem,vse}. 
There are also methods~\citep{SP-AEN,EDEZSL,TCN} that employ a separate shared space, distinct from both the image and attribute domains. To mitigate overfitting to known categories, these approaches often use pseudo-labeling techniques or operate in a transductive manner~\citep{EDEZSL}, accessing unlabeled images from new categories during the training phase.

More recently, embedding-based methods were extended to enable end-to-end training of the backbone~\citep{CC-ZSL,AMGML,AREN,Manigrasso2023FuzzyLV}. These methods, which include FLPN and will be denoted in the following as \textbf{end-to-end embedding-based} methods, are capable of identifying the essential areas of the image for class categorization and thus improving the embedding space during training~\citep{CC-ZSL,AREN,chen2022gndan}.
Previous research has exploited attributes to develop class-level representations using regularization methods~\citep{EDEZSL} or contrastive learning~\citep{CC-ZSL} to prevent overfitting of the seen classes.  Building on previous research~\cite{Du2022BoostingZL,Xu2022VGSEVS,yi2024prototype,chen2022gndan}, FLPN enhances its performance by combining attribute and class representations and leveraging the logical reasoning capability of a NeSy framework.

Compared to existing methods that employ a visual-semantic space, FLPN differs in the way it promotes alignment of the visual and semantic information. Techniques such as CoAR-ZSL \cite{Du2022BoostingZL} and APN\cite{Akata_2013_CVPR} define an auxiliary loss in the semantic space, after performing a visual-to-semantic mapping. Other techniques \citep{chen2021hsva} target the alignment between a visual and a semantic embedding space, where the latter is defined based on the text embedding of the attributes, a textual description of the class, or other sources of textual information. Our proposed NeSy approach, on the other hand, exploits structured knowledge (e.g., extracted from knowledge graphs) to regularize the learning process. On the one hand, it is a more flexible approach as the LTN knowledge base can encode different sources of information and constraints, but requires more tuning as it introduces additional hyper-parameters (such as the choice of grounding and the connective semantics). 

Previous approaches such as ProtoLTN
 \cite{Martone2022PROTOtypicalLT}, FLVN \cite{Manigrasso2023FuzzyLV} and CSNL \cite{Sikka2020ZeroShotLW} have successfully integrated symbolic knowledge with neural embeddings for ZSL, but there remain limitations in capturing attribute-level information and enforcing comprehensive logical constraints. FLPN addresses these gaps through several key innovations. FLPN introduces attribute-level prototypes in addition to class and macroclass prototypes, enabling finer recognition and better distinction between subtle visual differences in unseen classes. Furthermore, FLPN integrates a more diverse set of logical axioms within the LTN framework, including class hierarchies, exception handling, and similarity constraints. This comprehensive knowledge base provides stronger inductive biases and improves generalization to unseen classes by ensuring logical consistency and semantic coherence across different levels of abstraction. Additionally, FLPN employs specialized attention mechanisms tailored for both convolutional and transformer-based backbones, facilitating accurate attribute localization within images. By embedding prior knowledge directly into the feature representation and utilizing a unified embedding space, FLPN achieves state-of-the-art performance with reduced computational overhead compared to generative NeSy approaches. Lastly, FLPN is compatible with both CNN and transformer backbones, leveraging the advanced feature extraction capabilities and attention mechanisms of transformer models to achieve substantial performance gains on complex datasets.

\textbf{Generative techniques} utilize auxiliary models such as Generative Adversarial Networks (GANs) to create synthetic instances that represent unobserved classes by learning a conditional probability distribution for each class~\citep{TGMZ,LisGAN,cycle-CLSWGAN,CEGZSL, gao2024improving,zhang2022zero}. Feature generation models have recently been combined with embedding-based models in a contrastive framework~\citep{CEGZSL}. Generative approaches require prior knowledge of unseen classes to generate training data. In contrast, the FLPN training approach relies on choosing a set of seen classes and only requires knowledge of the complete class hierarchy during the training phase, facilitating the inclusion of new unseen classes in the training process.  FLPN is also compatible with generative techniques and can, in principle, integrate synthetic samples during training.

Finally, more recently transformers-based models, that is, techniques that integrate transformers as their backbone, have shown enhanced robustness compared to convolutional neural networks, or CNNs. According to Du et al. \cite{Du2022BoostingZL}, it is feasible not only to use global features for embedding spaces, but also to apply Visual Transformer (ViT) embedding patches as an attention mechanism to identify attribute-level features. Recent transformer-based models with dual attention pathways improve visual-semantic alignment, addressing the semantic gap in zero-shot learning \cite{cheng2023hybrid}. Incorporating topological structure enhances the classification of unseen categories \cite{chen2023integrating}, while linking sample clusters to class prototypes mitigates domain shift, reinforcing generalized zero-shot learning \cite{zhang2022zero}.
Both Hou et al. \cite{hou2024zeromamba} and Ran et al. \cite{ran2023zero} refine visual-semantic alignment by combining attention mechanisms with structured feature representations, leading to improved classification of unseen categories. Hou et al. employ a hybrid local-global attention strategy, while Ran et al. incorporate graph-based techniques for topology-aware feature learning. FLPN incorporates similar attention mechanisms, but trained within a NeSy formulation, achieving a comparable performance boost.

\subsection{NeSy AI for semantic image interpretation}

This research is part of a larger effort within the NeSy domain to incorporate symbolic information from the outset of training ~\citep{van2021modular}.
In particular, LTNs are a modular framework designed to incorporate  FOL constraints~\citep{LTN,Badreddine2023logLTNDF,yu2021survey}. They have been effectively applied in various computer vision tasks, such as object detection~\citep{donadello_semantic_image_interpretation,manigrasso2021faster}, ZSL~\citep{Martone2022PROTOtypicalLT,Manigrasso2023FuzzyLV} and visual reasoning tasks such as Sudoku puzzle classification~\citep{Morra2023DesigningLT}. Several studies highlight the potential to compensate for limited supervision by incorporating prior knowledge~\citep{Manigrasso2023FuzzyLV,donadello2019compensating,manigrasso2021faster}.

Within the LTN framework, grounding $\mathcal{G}$ is the process of assigning real-valued semantics to logical symbols. In this framework, individuals are represented as vectors in a high-dimensional space and each predicate symbol, denoted by $p \in \mathcal{P}$, is mapped to a function $\mathcal{G}(p) \rightarrow [0,1]$. An example is the \texttt{isOfClass} predicate commonly used in NeSy architectures, which measures the probability that a given term belongs to a specific class $c$~\citep{manigrasso2021faster, Martone2022PROTOtypicalLT}. Predicates are typically implemented using neural tensor networks (multi-layer perceptrons, or MLPs), but they can also be realized as measures of distance between terms (or better, between their vectorized groundings) and class-defining attribute matrices~\citep{Manigrasso2023FuzzyLV} or prototypes~\citep{Martone2022PROTOtypicalLT}. Complex expressions are constructed using a combination of logical connectives ($\land$, $\lor$, $\rightarrow$, $\neg$) and quantifiers ($\forall$, $\exists$), which are grounded through mathematical operations and neural network architectures. The training objective is established by defining a knowledge base $\mathcal{K}$ of the FOL axioms and then optimizing for the best satisfiability (sat) problem by maximizing the truth values of all formulas $\phi \in \mathcal{K}$ $
    \Theta^{*} = \text{argmax}_{\Theta} \left(\hat{\mathcal{G}}_{\theta}\left(\bigwedge_{\phi \in \mathcal{K}} \phi \right) - \lambda ||\Theta||_{2}^{2}\right)$.

\section{Methodology}
\label{sec:flpn}
\begin{figure*}[tb!]
\centering

    \includegraphics[width=\linewidth]{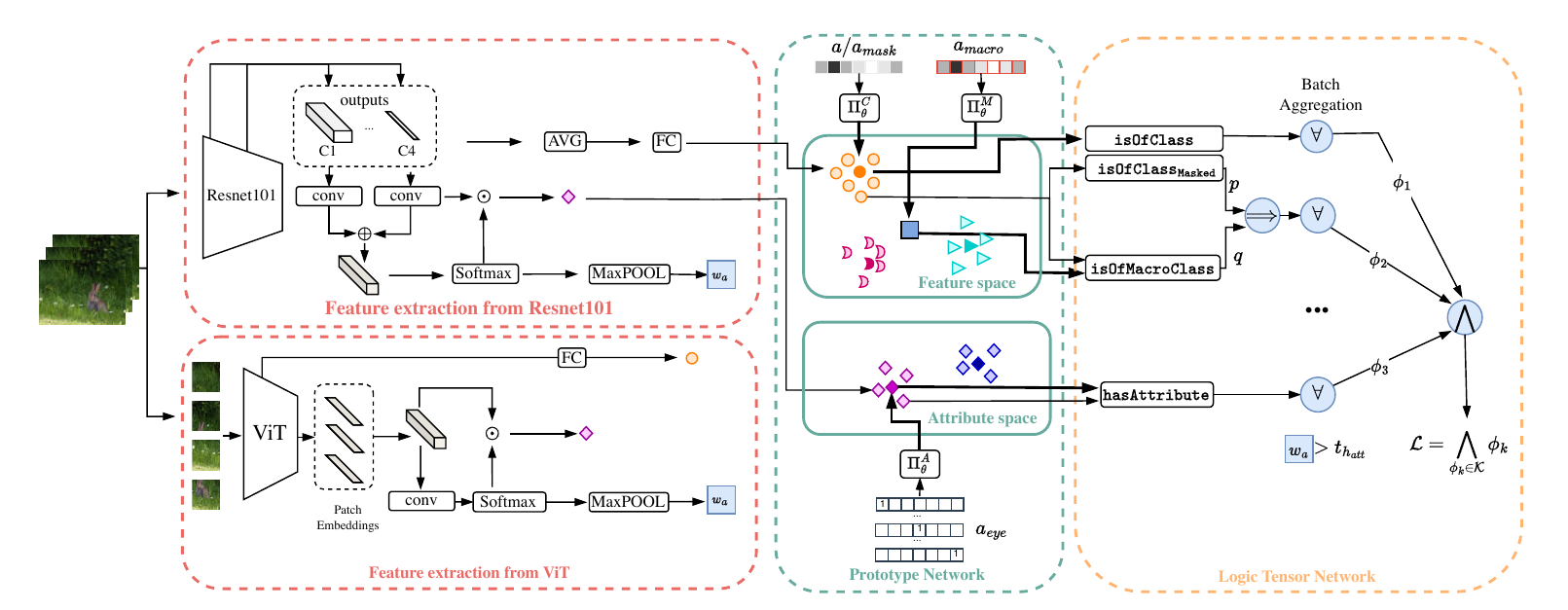}
\caption{The Fuzzy Logic Prototype Network (FLPN) architecture is composed of two main components: the feature extractor and the prototypical network. The feature extractor can be either a CNN (Resnet-101 in our experiments) or a Visual Transformer (ViT). The prototypical networks ($\CP$, $\MP$, $\AP$) integrate input images and the class ($a$), macroclass ($a_{macro}$), and attribute ($a_{eye}$) labels into a single embedding space, which in LTN terminology grounds the symbols in the embedding space. Building on this embedding space, various predicates, such as $\texttt{isOfClass}$, $\texttt{isOfClass}_{\text{masked}}$, and $\texttt{isOfMacro}$ are defined as class membership functions. In addition, the $\texttt{hasAttribute}$ predicate is used to detect specific attributes in images. These predicates constitute the foundational elements of the knowledge base, denoted $\KK$, within the LTN framework. The training objective (loss function) is formulated to optimize the satisfiability or truth value of $\KK$. The symbols $\oplus$ and $\odot$ denote the addition and multiplication operations, respectively.
}
\label{Figure:architecture}
\end{figure*}

The architecture of FLPN is depicted in Figure~\ref{Figure:architecture}. Its main components include the backbone, which is responsible for extracting the features from images (detailed in Section~\ref{Feature Extraction}), and the prototypical network (detailed in Section~\ref{Prototypes_mapping}), which generates prototypes for different classes, macroclasses, and attributes (as illustrated in Section~\ref{Prototypes_mapping}). The architecture was tested with two distinct backbones: a CNN (ResNet-101) and a transformer (ViT). Sections~\ref{ltn} and \ref{axioms} detail the process of integrating prototypes and images into a unified embedding space, which serves as the core of the knowledge base $\KK$, continually refined throughout the training phase~\citep{Badreddine2023logLTNDF}. Section~\ref{analysis} illustrates how optimizing the satisfiability of the grounded knowledge base relates to  losses commonly used in prototype and contrastive learning. 

\subsection{ZSL and GZSL settings}
\label{subsec:zsl_gzsl}

The ZSL task entails recognizing objects from previously unseen classes by exploiting
some form of auxiliary knowledge, usually attribute-based, and learning a mapping on seen
classes. GZSL extends the ZSL setting by assuming that both seen and unseen classes are present
at test time. Formally, the training dataset is defined as \( S = \{ (x_{n}, y_{n}) \mid n=1,\ldots,N \} \), with the training categories denoted by \( y_{n} \in Y^{tr} \). In the GZSL setting, a test image can be classified into seen or unseen classes, represented as \( Y^{ts+tu} \subset Y \), whereas in ZSL we assume that seen classes are not present at test time (\( Y^{tu} \subset Y \)).

Let us define \( C \) attribute vectors \( \{ a^{1}, a^{2}, a^{3}, \ldots, a^{C} \} \), where each vector \( a^{c} \in \mathbb{R}^{A} \) denotes attributes typically associated with a class \( c \in C \).
The entries in the attribute matrix \( a = \left[ a^{1}, \ldots, a^{C} \right] \) are non-binary: a zero value indicates an undefined attribute for the class, a negative value indicates that the attribute is not present in the class, and a positive value indicates the attribute is present with a relevance proportional to the value~\citep{awa2}. Similarly to previous methods~\citep{Du2022BoostingZL,Martone2022PROTOtypicalLT,dem}, we map the input images and the attribute matrix onto a \textit{common embedding space} by introducing functions $f_{\theta}$ and $g_{\theta}$, defined in Section~\ref{Feature Extraction}, to map images from their native domain, and a collection of functions (defined in Section~\ref{Prototypes_mapping}) that transform attribute vectors into \textit{prototypes} into the common embedding space.

\subsection{Image feature extraction}
\label{Feature Extraction}
Two feature extraction methods are presented, based either on a CNN or a ViT. These modules gather detailed image information, producing feature embeddings at the class, macroclass, and attribute levels to minimize distances to the representative prototypes.

\subsubsection{Global features representation}

Similarly to PROTO-LTN~\citep{Martone2022PROTOtypicalLT},  images are mapped into a shared embedding space in which both classes and macroclasses are represented through function:
\begin{align}
f_{\theta}: \mathbb{R}^{Ch \times W \times H} \rightarrow \mathbb{R}^{M}
\end{align}
where \( D = Ch \times W \times H \) is the input image domain, $W$, $H$ and $Ch$ are the image width, height and number of channels,  $M$ is the dimensionality of the feature space, and \( \theta \) are the model parameters.

\textbf{CNN-based}. The global features are extracted directly from the last layer of the Resnet-101 backbone.

\textbf{ViT-based}.
The images are divided into square patches and passed through the transformer encoder to extract the feature tensors. In contrast to CNN-based architectures, the class-level feature is represented by adding a learnable classification token [CLS] as described in~\citep{Du2022BoostingZL}.

\subsubsection{Attribute features representation}
An additional embedding function converts each image into suitable attribute embeddings through an attention mechanism:
\begin{align}
g_{\theta}: \mathbb{R}^{Ch \times W \times H} \rightarrow \mathbb{R}^{A \times M }
\end{align}
where \( D = Ch \times W \times H \) is the input image space, \( M \) the dimensionality of the shared feature space, \( A \) the number of attributes, and \( \theta \) a set of trainable weights.  Similarly to previous work \cite{Du2022BoostingZL}, $g_{\theta}$ incorporates an attention mechanism on top of the shared feature extraction module used by $f_{\theta}$, so that the backbone parameters are shared among the two functions.

\textbf{CNN-based module}. We added four convolutional layers after each stage to generate feature tensors. These tensors are combined to form an attention map, which is normalized using the Softmax function. Each feature map in the attention map acts as a soft mask to highlight probable attribute locations. Attribute-specific features are extracted through bilinear pooling between the attention map and the image feature tensor, involving pixel-wise multiplication and average pooling to localize attributes, as detailed in \cite{Du2022BoostingZL}.

\textbf{ViT-based module}. An attention-based attribute localization method, analogously to the CNN-based architecture, is defined. Features extracted from the backbone are passed through a convolutional layer to generate the feature tensor. A soft mask for each attribute localization is created using a max-pooling function. Each attribute feature is then bi-linearly pooled with the image feature to obtain the attribute-level feature.

\subsubsection{Attribute features weight}
In GZSL and ZSL, attributes are often defined at the class level, and individual annotations for each image are not available.  An attribute weight vector $W_a \in \mathbb{R}^{A}$ acts as a data-driven gating mechanism to identify attributes that are not visually present, despite being associated with the class, by applying a threshold $th_{att}$. Although the weight vector $W_a$ may seem conceptually redundant, it substantially reduces the computational cost of computing $\KK$. Further details on this efficiency gain will be elaborated later in the paper.

\subsection{Prototypes for semantic mapping}
\label{Prototypes_mapping}
Each entity, be it a class, a macroclass, or an attribute, is represented by a prototype embedded in a shared embedding space of size $M$. These prototypes serve as reference points to classify the input images by comparing them with their respective embeddings.
Different parameterized functions, collectively denoted as $\Pi_{\theta}$  are employed to learn the prototypes for classes, macroclasses, and attributes. We employ here the term macroclass to indicate categorization at different semantic levels, as induced by a hypernym-hyponym relationship: in our datasets, we consider a ``tiger'' to be a class and ``feline'' to be the corresponding macroclass. Different groundings of the class-membership function may be empirically more effective depending on the type of categorization and the desired level of mutual exclusivity.

\begin{itemize}
 \item \textbf{Class Prototypes}: $\CP$ transforms the attribute matrix (denoted by $a$, of dimension $C \times A$,  where $C$ is   the number of classes and $A$ the number of attributes) into Class Prototypes in the shared space $M$.

 \item \textbf{Macroclass Prototypes}: similarly to class prototypes, $\MP$ maps the attribute matrix $a_{macro}$ (dimension $C_{\text{macro}} \times A$, where $C_{\text{macro}}$ is the number of macroclasses) into Macroclass Prototypes within the shared space $M$. These prototypes merge the C classes into a smaller set of $C_{\text{macro}}$ macroclasses based on attributes similarities. We employ a trainable matrix $a_{macro} \in \mathbb{R}^{C_{macro} \times A}$ to identify the attributes relevant to each macroclass, as they are not available in the ground truth.

 \item \textbf{Attribute Prototypes}: individual attributes are mapped into a shared space by a mapping function $\AP$ which takes as input the eye matrix, which represents the one-hot encoded attribute labels.
\end{itemize}

The initial layers of the network modules that implement the three prototype functions ($\CP$, $\MP$, and $\AP$) share the same weights.

\subsection{Logic Tensor Network}
\label{ltn}

The LTN reformulates the learning objective by optimizing the satisfiability of a knowledge base ($ \KK$). For each training batch, $ \KK$ is calculated starting from axioms based on labeled training instances $(\axclass{})$ and prior domain knowledge $(\axmacro{}, \phi_{\text{masked}}, \axattr{}, \axsimil{}, \axdissimil{})$. A detailed overview of the $\KK$ construction is provided in the remainder of this section.

The maximum satisfiability loss is then defined based on the aggregation of all axioms as follows:
\begin{equation}
        \mathcal{L}^{\text{ep}} = 1 -\left(\bigwedge_{\phi \in \mathcal{K}} \phi \right)  = 1 -\GG(\phi)
        \label{eq:loss}
\end{equation}

\subsubsection{Variables}
\label{sec:groundings}

The following variables are defined:
$\mathcal{G}(x)  = \mathbb{R}^{Ch\times H \times W}$;
$\mathcal{G}(l) = \mathbb{N}^{C}$;
$
\mathcal{G}(l_m) = \mathbb{N}^{C_{macro}}$;
$
\mathcal{G}(l_a) = \mathbb{N}^{C \times A}$;
$\mathcal{G}(w_{a}) \in \mathbb{R}^{A}$;
$
\mathcal{G}(a) = \mathcal{G}(a_{\text{mask}}) = \mathbb{R}^{C \times A} $;
and $\mathcal{G}(a_{\text{macro}}) = \mathbb{R}^{C_{macro} \times A}$. Here, $x$ denotes an RGB image with size $Ch \times H \times W$, $l$, $l_m$ and $l_a$ represent class, macroclass, and attribute labels, $C$,$C_{\text{macro}}$ and $A$ are the total number of classes, macroclasses, and attributes, and $M$ is the dimensionality of the embedding space.

\subsubsection{Predicates and functions}

We define four key predicates. First, the $\texttt{isOfClass}(x, l)$ predicate evaluates the probability that an image $x$ belongs to class $l$. This is achieved by computing the cosine similarity between the global features of the image, extracted by $f_{\theta}$, and the class prototype computed by $\CP(a_l)$, where $a_l$ denotes the attribute vector for class $l$. The probability is computed using a Softmax function as:
\vspace{-10pt}
\begin{equation}
 p(x, l) = \frac{\exp \left( \delta \cos(f_\theta(x), \CP(a_l)) \right)} {\sum_{s=1}^S \exp \left(\delta \cos(f_\theta(x), \CP(a_s)) \right)} \label{eq:prob}
\end{equation}
where $\delta$ is a constant value that acts as a temperature parameter.

The prediction score for an instance $x$ with respect to class $c$ (that is, the degree of truthiness of the predicate $\texttt{isOfClass}$ for a specific input $x$ and class $c$) is computed by taking the dot product of the one-hot encoded class label $l_c^T$ (where $T$ denotes transpose) with the output of the function $p$ as:

\vspace{-10pt}
\begin{equation}
\mathcal{G}(\texttt{isOfClass})(x, l_c) = l_c^T p(x, l_c)
\label{eq:isOfClass_predicate}
\end{equation}

We further define a masked version $\texttt{isOfClass}_{\text{masked}}$. Its grounding is the same as that of the $\texttt{isOfClass}$ predicate, except that a masked attribute vector $a_c^{\text{masked}}$ is used. This predicate was specifically introduced, as in FLVN~\citep{Manigrasso2023FuzzyLV}, to address potential discrepancies between image-level and class-level attributes (for example, a front-facing photo of a bird might not show its tail). The masked attribute vector $a_{mask} \in \mathbb{R}^{C \times A}$ has the same content of the attribute vector $a_c$, except that $k$ elements are randomly masked (i.e., set to 0). Unlike FLVN~\citep{Manigrasso2023FuzzyLV}, FLPN randomly drops only attributes with scores in $a_c$ lower than the class mean, that is attributes which are less frequently observed within the class.

The $\texttt{isOfMacro}(x, l_m)$ predicate computes the likelihood that an image $x$ belongs to the macroclass $m$.
It is grounded as a function of the distance between the embedding of the image $x$ and the prototype embedding representing the macroclass $l_m$:
\begin{equation}
\GG(\texttt{isOfMacro}): x,l_{m} =   \mathrm{e}^{-\alpha \bigg(\text{d}\big(f_{\theta}(x), \MP(a_{m})\big)\bigg)}
\label{eq:gt_is_macro}
\end{equation}
where $x$ is an input image, $l_m$ is the macroclass label, $a_m$ is the corresponding attribute matrix, $\alpha$ is a hyperparameter and $\text{d}(\cdot, \cdot)$ is a measure of distance. In our experiments, $\text{d}(\cdot, \cdot)$ is the cosine distance (1 minus the cosine similarity), while PROTO-LTN~\citep{Martone2022PROTOtypicalLT} and DEM~\citep{dem} used the Euclidean distance. $\GG(\texttt{isOfMacro})$ takes the value of $1$ when the distance from the macroclass prototype is zero.

The $\texttt{hasAttribute}$ predicate computes the probability that an image $x$ is associated with the attribute label $l_a$. It is again computed as a function of the cosine distance between an embedding and an attribute prototype:
\vspace{-10pt}
\begin{equation}
\GG(\texttt{hasAttribute}): x,l_{a} =   l_{a}^T  \cdot  \mathrm{e}^{-\beta \bigg( \text{d}\big(g_{\theta}(x), \AP(a_{eye})\big)\bigg)}
\label{eq:gt_hasAttribute}
\end{equation}

Experimentally, we observed better results when using different distance measures for the predicates $\texttt{isOfClass}$, $\texttt{isOfMacro}$, and $\texttt{hasAttribute}$. We postulate that the effectiveness of the $\texttt{isOfClass}$ grounding (\autoref{eq:isOfClass_predicate}) stems from promoting class exclusivity, while the grounding of the predicates $\texttt{isOfMacro}$ and $\texttt{hasAttribute}$ allows greater overlap of class attributes.

\subsection{Knowledge base axioms}
\label{axioms}
\textbf{Learning from labeled examples. }
\label{sec:learning_from_labeled_examples} Axiom $\axclass{}$ promotes predictions that are consistent with the data labels (in natural language, ``this image is a zebra''):
\vspace{-10pt}
\begin{equation}
\axclass{} = \forall \text{Diag}(x,l_{c}). \texttt{isOfClass}(x,l_{c}) \label{eq:isOfClass} \footnote{Diagonal Quantification quantifies over pairs of instances, e.g., images and their labels. A more formal definition can be found in~\citep{LTN}.}
\end{equation}

Prior knowledge about class hierarchies (e.g., ``zebras belong to the class of ungulates'') is encoded by axiom $\axmacro{}$:
\begin{equation}
\begin{split}
\axmacro{} = \forall \text{Diag}(x,l_c,l^c_{macro}). \ & \texttt{isOfClass}(x,l_c) \\
& \implies \texttt{isOfMacro}(x,l^c_{macro})
\end{split}
\label{eq:isOfMacro}
\end{equation}
where $l^c_{macro}$ is the macroclass associated with class $c$.

\textbf{Learning with exceptions}.
\label{sec:attribute_incompleteness} The axiom $\phi_{\text{masked}}$ captures the notion that not all class attributes are present in every sample (e.g.,``there exists a zebra that is not agile'' or ``there is an image of a bird in which the tail is not visible''). The axiom is based on the $\texttt{isOfClass}_{\text{masked}}$ predicate:
\vspace{-10pt}
\begin{equation}
\phi_{\text{masked}} = \forall l^{\text{seen}}
( \exists x. \texttt{isOfClass}_{\text{masked}}(x,l^{\text{seen}}))
\label{eq:isOfClassmasked}
\end{equation}
where $l^{\text{seen}}$ denotes the list of seen classes. Let us remind that the $\texttt{isOfClass}_{\text{masked}}$ predicate randomly masks $k$ attributes for each class from the attribute vector when computing the class prototypes. Specifically, we calculate the average of all non-zero entries per class in the attribute vector $a_c$ and select $k$ attributes with values below this average to be masked. Other strategies are possible, e.g., random selection regardless of the attribute matrix $a$, but we empirically found them to be less effective.

\textbf{Learning with attention.} \label{sec:Learning_with:attention} The axiom $\axattr{}$ encourages images with similar attributes to reside close together within the embedding space, regardless of their associated class:
\vspace{-10pt}
\begin{equation}
\begin{split}
\axattr{} &= \forall \text{Diag}(x, w_{a}, l^c_a):  w_a[l^c_a] > th_{att}.   \texttt{hasAttribute}(x, l^c_a)
\end{split}
\label{eq:hasAttribute}
\end{equation}
where $x$ represents an input image, $l^c_a$ identifies the non-zero attributes linked to its class, and $w_a$ stands for the attribute weight vector. For greater computational efficiency, we focus on attributes $l^c_a$ that are both relevant to the class (based on the attribute matrix $a$) and likely present in the image $x$, by considering only attributes $l$ with weights $w_a^l$ exceeding a threshold $th_{att}$. In practice, since the knowledge base $\KK$ is updated every training batch, we leverage current attention values to enforce this criterion.

\textbf{Learning better feature representations.} Axioms $\axsimil{}$ and $\axdissimil{}$ encourage image embeddings of similar images to be clustered together within the embedding space, while maintaining the separation between distinct classes. Specifically, to group similar image representations belonging to the same class, we introduce the axiom $\axsimil{}$:
\vspace{-10pt}
\begin{equation}
\begin{aligned}
\axsimil{} &= \forall l^{\text{seen}},\forall x,\forall y:
\text{cos}(f_\theta(x),f_\theta(y))>th_{sg}. \\& \texttt{isOfClass}(x,l^{\text{seen}}) \Leftrightarrow \texttt{isOfClass}(y,l^{\text{seen}})
\end{aligned}
\label{eq:sameclass}
\end{equation}
where $f_\theta(x)$ and $f_\theta(y)$ represent the visual embeddings of images $x$ and $y$.
The axiom simply states that when two images $x$ and $y$ are similar, based on their respective cosine similarity $\text{cos}(f_\theta(x),f_\theta(y))$, then they should be classified as belonging to the same class. To keep the size of $\KK$ within manageable limits, we restrict the axiom to image pairs $(x, y)$ with cosine similarity exceeding a threshold $th_{sg}$.

Likewise, axiom $\axdissimil{}$ increases the distance between images belonging to different classes, based on the cosine similarity between their respective features:
\vspace{-10pt}
\begin{equation}
    \begin{aligned}
\axdissimil{} &= \forall l^{\text{seen}},\forall x,\forall y:   \text{cos}(f_\theta(x),f_\theta(y))<th_{sl}. \\&  \neg\big(\texttt{isOfClass}(x,l^{\text{seen}}) \land \texttt{isOfClass}(y,l^{\text{seen}})\big)
\end{aligned}
\label{eq:notsameclass}
\end{equation}

Again, $\KK$ includes only pairs of images with cosine similarity lower than a threshold $th_{sl}$.

\subsubsection{Grounding logical connectives and aggregators}
\label{subsec:grounding}

 We experimented with two different operator semantics, denoted in the following as LTN~\citep{LTN} and logLTN~\citep{Badreddine2023logLTNDF}, obtaining two versions of our architecture which we denote for brevity FLPN and Log-FLPN. LTN operates in the probability space and logLTN in the log probability space~\citep{Badreddine2023logLTNDF}. Table~\ref{tab:connective} outlines the main differences between the logical connectives and aggregation operators used in logLTN versus the standard LTN formulation. Further details on the roles of $\exists$ and $\forall$ can be found in~\cite{LTN,Badreddine2023logLTNDF}.

\begin{table*}[tb!]
 \centering
 \begin{tabular}{c|c|c}
 \textbf{Connective} & \textbf{LTN} & \textbf{logLTN} \\ \hline
 $\neg$ & $1 - \sigma(a)$ & $ \log(1-e^{\log(\sigma(a))})$ \\

 $\land$ & $\sigma(a) * \sigma(b)$ & $\log(\sigma(a)) + \log(\sigma(b))$ \\

 $\lor$ & $\sigma(a) + \sigma(b) - \sigma(a)* \sigma(b)$ & ${\frac{1} {\alpha}} C+\log \left( {\frac{1} {n}\sum_{i=1}^{n} e^{\log(\alpha \sigma(a_{i})) - C}} \right)  $ \\

 $\rightarrow$ & $1 - \sigma(a) + \sigma(a)* \sigma(b)$ & $a \rightarrow b \Leftrightarrow \neg a \lor b$ \\

 $\forall$ & $1-\left( \frac{1} {n} \sum_{i=1}^{n} \left( 1-\sigma(a)_i \right)^{p} \right)^{\frac{1} {p}}$ & $\frac{1} {n} \sum_{i=1}^{n} \log(\sigma(a)_i) $ \\

 $\exists$ &
 $\left( {\frac{1} {n}} \sum_{i=1}^{n} \sigma(a)_i^{p} \right)^{\frac{1} {p}}$ &
 ${\frac{1} {\alpha}} C+\log \left( {\frac{1} {n}\sum_{i=1}^{n} e^{\log(\alpha \sigma(a_{i})) - C}} \right)  $ \\

$\Leftrightarrow$ & $(1 - \sigma(a) + \sigma(a)*\sigma(b))*(1 - \sigma(b) + \sigma(a)*\sigma(b))$ & $(\neg a \lor b) \land (\neg b \lor a)$ \\

 \end{tabular}
 \caption{Grounding of logical connectives and aggregators ($p \geq 1$). In the existential quantifier, $C = \max(\alpha \log(\sigma(x)_{i}))$. $\sigma$ is the activation function. }
\label{tab:connective}
\end{table*}

\subsubsection{Querying the knowledge base}
During inference, the class with the maximum score is chosen as the predicted class. The Calibrated Stacking method, with parameter $\delta$ was used to reduce the bias towards classes seen at training time by lowering their scores~\citep{Chen2022MSDNMS,Wang2023GeneralizedZA}

\subsection{Relationship to contrastive learning}
\label{analysis}

After grounding the axioms, interesting connections can be drawn between their mathematical formulation and loss functions commonly used in deep learning. In the following, unless otherwise noted, we will use the logLTN grounding for the logical connectives.

Let us consider axiom $\axattr{}$ defined in Eq. \ref{eq:hasAttribute}. After grounding the universal quantifier as defined in Table \ref{tab:connective} and substituting the grounding of the $\texttt{hasAttribute}$ predicate defined in Eq. \ref{eq:gt_hasAttribute}, the satisfiability of the axiom is computed as follows:

\begin{equation}
    \axattr{} = \sum_{i=1}^{N_b}\sum_{j=1}^{\Tilde{K}_i}{\bigg( \log (\mathrm{e}^{-\beta  \text{d}(g_{\theta}(x), \AP(a_{eye}))})\Bigg)}
    \label{eq:axattr_grounded}
\end{equation}
where $N_b$ is the number of examples in the current batch, $\Tilde{K}_i$ the list of attributes that are relevant for the specific image, and $d(\cdot)$ is a distance function. This loss is, up to a scaling constant, equivalent to existing losses for learning a deep embedding model for zero-shot learning \cite{dem,Martone2022PROTOtypicalLT}.

Following the same approach, for the  $\axclass
{}$ axiom defined in Eq. \ref{eq:isOfClass}, we obtain the following grounded formulation:
\begin{equation}
    \axclass{} = \sum_{i=1}^{N_b}{ \log \frac{\exp \left( \delta \cos(f_\theta(x), \CP(a_l)) \right)} {\sum_{s=1}^S \exp \left(\delta \cos(f_\theta(x), \CP(a_s)) \right)}}
    \label{eq:axclass_grounded}
\end{equation}

Maximizing Eq. \ref{eq:axclass_grounded} corresponds thus to minimizing the loss defined in the original prototype network \cite{prototypical}. Moreover, as suggested in \cite{liprototypical}, Eq. \ref{eq:axclass_grounded} can be seen as a variant of the InfoNCE loss used by many popular contrastive learning methods \cite{he2020momentum}:

\begin{equation}
    \mathcal{L}_{\text{InfoNCE}} = \sum_{i=1}^{N_b} -\log \frac{\exp(\mathbf{z}_i \cdot \mathbf{z}'_i / \tau)}{\sum_{j=1}^{R} \exp(\mathbf{z}_i \cdot \mathbf{z}'_j / \tau)}
\end{equation}

where $N_b$ is the number of samples in a batch, containing one positive embedding and R negative embeddings. As suggested in \cite{liprototypical}, the two formulations coincide if $\delta$ replaces the temperature parameter $\tau$, the augmented positive instance $\mathbf{z}'_i$ is replaced by the class prototype $\CP(a_l)$ and the augmented negative instances $\mathbf{z}'_j$ are replaced by the other class prototypes $\CP(a_s)$ . The view that the $\axclass{}$ introduces a contrastive element to the training procedure is reinforced by the way in which the training batch is sampled (detailed in Section \ref{sec:experiments}), ensuring a certain ratio of positive to negative examples, which we experimentally found to be crucial in experiments with smaller batch sizes (i.e., when using a ViT backbone).  This observation suggests a potential connection with the performance of our NeSy approach and the batch size, which we leave for future work.

Similarly, $\axsimil{}$ and $\axdissimil{}$ introduce a form of contrastive learning that encourages similar images to cluster around their corresponding prototype. Let us consider for instance  axiom $\axdissimil{}$ defined in Eq. \ref{eq:notsameclass}, and define
$P = \{ (x_i, y_i) \mid i = 1, 2, \ldots, K \}$ as the list of dissimilar image pairs selected by the guarded universal quantifier. When ground, $\axdissimil{}$ can be expressed as:

\begin{equation}
\begin{aligned}
\axdissimil{} = \sum_{l=1}^L \sum_{i=1}^P & \log \bigg(1 - \frac{\exp \left( \delta \cos(f_\theta(x_i), \CP(a_l)) \right)} {\sum_{s=1}^S \exp \left(\delta \cos(f_\theta(x_i), \CP(a_s)) \right)} \bigg) \\
& \cdot \log \frac{\exp \left( \delta \cos(f_\theta(y_i), \CP(a_l)) \right)} {\sum_{s=1}^S \exp \left(\delta \cos(f_\theta(y_i), \CP(a_s)) \right)}
\end{aligned}
\label{eq:notsameclass_grounded}
\end{equation}

which encourages a prototype space in which, for each class label $l$, the corresponding prototype $\CP(a_l)$ is either pushed towards $x_i$ or $y_i$, but not both.

It must be stressed that the exact mathematical formulation also depends on the chosen grounding (or semantics) of the logical connectives. When choosing the alternative LTN formulation in Table \ref{tab:connective}, the universal aggregator becomes the p-mean error $1-\left( \frac{1} {n} \sum_{i=1}^{n} \left( 1-\sigma(a)_i \right)^{p_\forall} \right)^{\frac{1} {p_\forall}}$, with $p_\forall=2$ in our experiments, which replaces the outer sum in Eqs. \ref{eq:axattr_grounded}, \ref{eq:axclass_grounded}, and \ref{eq:notsameclass_grounded}.
Mathematically, with $p_\forall > 1$ the loss can no longer be decomposed into the sum of independent components, one for each training instance. This adjustable hyper-parameter places more or less emphasis on outliers (see \cite{LTN} and \cite{Badreddine2023logLTNDF} for additional discussions on the role of this hyper-parameter, and Section \ref{sec:experiments} for the exact scheduling employed in our experiments). Furthermore, the satisfiability of the axiom the training convergence depends on the batch size and sampling strategy.

\section{Experiments}
\label{sec:experiments}
In this section, we present the datasets used in our experiments, the knowledge base adopted for each dataset, and the list of hyperparameters.

\textbf{Dataset}.The experiments were performed on AWA2~\citep{awa2}, CUB~\citep{CUB}, and SUN~\citep{SUN} benchmarks. The GZSL evaluation metrics followed the standards set by previous research~\citep{awa2}.

\textbf{Knowledge base}. The composition of the knowledge base $\KK$ is elaborated in Section~\ref{sec:learning_from_labeled_examples}. For constructing the class hierarchy, we associate classes from the AWA2 and CUB dataset with their respective synsets in WordNet, following the approach used in ~\citep{Sikka2020ZeroShotLW,Manigrasso2023FuzzyLV}.

Following the approach outlined in previous works ~\cite{Sikka2020ZeroShotLW, Manigrasso2023FuzzyLV}, we constructed a semantic hierarchy by grouping the classes from the AWA2 and CUB datasets into 9 and 49 macroclasses, respectively. Specifically, the macroclasses for the AWA2 and CUB datasets were formed by finding the root synset whose subtree includes the dataset classes, then assigning its direct descendants (via the WordNet hierarchy) as classes. It was not possible to define macroclasses for the SUN dataset, and hence the related axiom was excluded from the knowledge base.

For all experiments, within the predicate $\texttt{isofClass}_{masked}$, 15 attributes ($K=15$) were omitted at random. A consistent threshold value of $th_{att}=9.0$ was applied in all experiments, with the value of $\delta$ uniformly set to 50 for every dataset. The thresholds $th_{sg}$ and $th_{sl}$ were set at 0.5 and 0.3 for AWA2 and to 0.7 and 0.6 for CUB and SUN, respectively. To manage outliers, the initial settings for the parameters of the aggregation function in the standard implementation of LTN were set at $p_{\exists}=2$ and $p_{\forall}=2$ (see LTN in Table~\ref{tab:table_differences2}). These parameters were increased by 2 after each 4 epoch for all datasets, adhering to the protocol recommended in~\cite{LTN}.

\textbf{Hyperparameter selection}. The embedding functions $f_{\theta}$ and $g_{\theta}$ convert an image into a vector in a 2048-dimensional space for ResNet and a 1024-dimensional space for ViT. We trained the model end-to-end, using the SGD optimizer with an initial learning rate of $1e-3$. During the fine-tuning phase, the learning rate was subsequently decreased by a factor of $1e-7\iota$, where $\iota=0.8^{epoch//10}$, for a total of 300 epochs for both the ResNet-101 and ViT models. When training the ResNet-101 backbone, samples were chosen at random from a batch of 64 samples, while for ViT, training batches consisted of a ratio of 4 to 8 for positive and negative examples. We set the scaling factor $\gamma$ to 0.7 at inference time to calibrate the scores for the seen classes. As data augmentation, we used a random flip with 0.5 chance and a random crop in each experiment. Using the LTNtorch library~\citep{LTNtorch}, we developed the architectures in PyTorch 2.1.1 and trained it on a single Nvidia 3090 Ti GPU. Each experiment was repeated three times to report the mean and standard deviation.

\section{Results}
\label{sec:results}

\begin{table*}[tb!]
\centering
\resizebox{\textwidth}{!}{%
\begin{tabular}{c|cccc|cccc|cccc}
\hline
&\multicolumn{4}{c}{AWA2}&\multicolumn{4}{c}{CUB} &\multicolumn{4}{c}{SUN} \\
\hline
Model&T1&U&S&H&T1&U&S&H&T1&U&S&H \\

\hline
\multicolumn{13}{c}{Embedding-based methods} \\
\hline
Proto-LTN~\citep{Martone2022PROTOtypicalLT}$\dagger$&67.6 &32.0 &83.7 &46.2&48.8 &20.8 &54.3 &30.0 &60.4 &20.4 &36.8 &26.2 \\
DEM~\citep{dem}&67.1&30.5&86.4&45.1&51.7&19.6&57.9&29.2 &61.9&20.5&34.3&25.6 \\
VSE~\citep{vse}&\textbf{84.4}&45.6&\textbf{88.7}&60.2 &71.9&39.5&68.9&50.2&-&-&-&-\\

TCN~\citep{TCN}&71.2&61.2&65.8&63.4&59.5 &52.6&52.0&52.3&61.5&31.2&37.3&34.0\\
\hline
\multicolumn{13}{c}{End-to-end Embedding-based methods (CNN backbone)} \\
\hline
AMGML~\citep{AMGML}&71.7 &56.0&74.6&64.0&70.0 &58.2&55.7&56.9&59.7 &42.0&35.1&38.3 \\
CC-ZSL~\citep{CC-ZSL}&68.8 &62.2&83.1&71.1&74.3&66.1&73.2&69.5&62.4& 44.4&36.9&40.3 \\
CoAR-ZSL~\citep{Du2022BoostingZL} & 74.1 & 68.1 & 79.1 & 73.2 & \underline{79.2} & 70.9 & 77.3 & 74.0 & 66.7 & 50.6 & 38.0 & 43.4 \\
FLVN $\dagger$ &69.8&65.8&82.3&73.1& 71.2 & 62.6 & 83.1 & 71.5 &61.7&48.4&32.7&39.0\\
&\textpm 0.8 &\textpm 0.9&\textpm 0.3&\textpm 0.6& \textpm 0.2 & \textpm 0.5 & \textpm 0.3 & \textpm 0.2 &\textpm0.18&\textpm0.57&\textpm0.2&\textpm 0.1\\
&(71.0)&\textbf({67.1})&(82.8)&(74.1)&(71.4)& (63.2)& (\textbf{83.4}) & (71.7)&(61.9)&(48.9)&(32.9)& (39.1) \\

PRZSL \citep{yi2024prototype} & 73.6 & 65.8 & 77.8 & 71.3 & 77.1 & 68.8 & 63.7 & 66.2 & 64.2 & 53.6 & 37.7 & \underline{44.3} \\
AS-ZSL \citep{zhou2023attribute} & 68.9 & 66.5 & 78.38 & 71.9 & 78.5 & 65.8 & 78.2 & 71.5 & 62.2 & 39.7 & 37.2 & 38.3 \\

HRT \citep{cheng2023hybrid} & 67.3 & \textbf{78.7} & 58.9 & 67.4 & 71.7 & 63.5 & 62.1 & 62.8 & 63.9  & 26.9 & \textbf{53.2} &35.7 \\

CSDR \citep{gao2024improving} & 76.7 & 68.0 & \underline{82.7} & 74.6 & 80.3 & \textbf{73.1} & 78.9 & \textbf{75.9} & 64.5 & 43.9 &    \underline{44.6} & 44.2 \\

CTPM \citep{chen2023integrating} & 74.1 & 67.2 & 79.6 & 72.9 & 73.4 & 64.7 & 73.2 & 68.7 & 66.0 & 44.0 & 43.2 & 43.6 \\

MSDN\citep{chen2022msdn} & 70.1 & 62.0 & 74.5 & 67.7 & 76.1 & 68.7 & 67.5 & 68.1 & 65.8 & 52.2 & 34.2 & 41.3 \\

GNDAN\citep{chen2022gndan} & 71.0 & 60.2 & 80.0 & 69.0 & 75.1 & 69.2 & 69.6 & 69.4 & 65.3 & 50.0 & 34.7 & 41.0 \\
\hline

\textbf{FLPN}$\dagger$ & 71.43 & 67.80 & 85.93 & 74.53 & 72.73 & 71.00 & 78.50 & 74.53 & 66.73 & 53.53 & 38.50 & {44.83} \\
& \textpm 3.44 & \textpm 2.99 & \textpm 2.28 & \textpm 0.91 & \textpm  0.48 & \textpm 0.65 & \textpm 2.76 & \textpm 1.21 & \textpm 0.50 & \textpm 0.56 & \textpm 0.36 & \textpm 0.17 \\
& (74.87) & (\underline{70.79}) & (\underline{88.21}) & (\textbf{76.61}) & (75.01) & (71.65) & (\underline{81.26}) & (\underline{75.75}) & (\underline{67.23}) & (\underline{54.09}) & (38.86) & (\textbf{45.00}) \\

\textbf{Log-FLPN}$\dagger$ & 66.53 & 64.17 & 82.00 & 71.90 & 73.87 & 71.93 & 73.67 & 72.73 & 61.60 & 52.77 & 37.93 & 44.10 \\
& \textpm 4.97 & \textpm 5.15 & \textpm 1.50 & \textpm 3.18 & \textpm 0.42 & \textpm 0.54 & \textpm 1.64 & \textpm 0.62 & \textpm 7.64 & \textpm 2.18 & \textpm 0.66 & \textpm 0.33 \\
& (71.50) & (69.32) & (83.50) & (\underline{75.08}) & (74.29) & (\underline{72.48}) & (75.31) & (73.35) & (\textbf{69.24}) & (\textbf{54.94}) & (38.59) & (44.43) \\
\hline
\multicolumn{13}{c}{End-to-end embedding-based methods (ViT backbone)} \\
\hline
CoAR-ZSL~\citep{Du2022BoostingZL} &78.7 & 76.5 & \underline{88.2} & 82.0 & \textbf{79.9} & 72.5 & 76.3 & \underline{74.4} & \textbf{79.4} & \textbf{68.3} & \underline{55.0} & \textbf{61.0} \\
ZeroMamba~\citep{hou2024zeromamba} & 80.0 & 72.1 & 76.4 & 74.2 & 71.9 & 67.9 & \textbf{87.6} & \textbf{76.5} & 72.4 & 56.5 & 41.4 & 47.7 \\

ZSLViT~\citep{ran2023zero} & \textbf{90.1} & 52.4 & 64.3 & 57.6 & 56.9 & 64.6 & 52.8 & 43.8 & 57.4 & 43.2 & 53.2 & 47.3 \\

TransZero++\citep{chen2022transzero++} & 78.3 & 67.5 & 73.6 & 70.4 & 72.6 & 64.6 & \underline{82.7} & 72.5 & 67.6 & 48.6 & 37.8 & 42.5 \\

\hline

\textbf{FLPN}$\dagger$ & 80.97 & 80.73 & 83.17 & 81.87 & 76.43 & 70.10 & 70.30 & 70.00 & 74.77 & 58.60 & 53.90 & 56.13\\
& \textpm 1.97 & \textpm 1.82 & \textpm 3.24 & \textpm 2.19 & \textpm 1.16 & \textpm 3.59 & \textpm 5.03 & \textpm 2.14 & \textpm 0.56 & \textpm 1.79 & \textpm 1.21 & \textpm 0.88\\
& (\underline{82.94}) & (\textbf{82.55}) & (86.41) & (\underline{84.06}) & (77.59) & (\textbf{73.69}) & (75.33) & (72.14) & (\underline{75.32}) & (\underline{60.39}) & (\textbf{55.11}) & (\underline{57.01})\\
\hline

\textbf{Log-FLPN}$\dagger$ & 78.23 & 77.10 & 86.23 & 81.40 & 75.20 & 69.63 & 63.10 & 65.93 & 73.37 & 58.03 & 51.43 & 54.53\\
& \textpm 2.60 & \textpm 2.12 & \textpm 5.11 & \textpm 3.34 & \textpm 2.86 & \textpm 3.65 & \textpm 6.64 & \textpm 3.64 & \textpm 0.45 & \textpm 1.11 & \textpm 0.46 & \textpm 0.69\\
& (80.83) & (\underline{79.22}) & (\textbf{91.34}) & (\textbf{84.74}) & (\underline{78.06}) & (\underline{73.29}) & (69.74) & (69.57) & (73.82) & (59.15) & (51.90) & (55.23)\\

\hline
\multicolumn{13}{c}{Generative methods} \\
\hline

Cycle-CLSWGAN~\citep{cycle-CLSWGAN}&-&-&-&-&58.4&45.7&61.0&52.3&60.0&49.4&33.6&40.0 \\
TGMZ~\citep{TGMZ}& \underline{78.4} &64.1& 77.3&70.1&66.1 &60.3&56.8&58.5&-&-&-&-\\
CEGZSL~\citep{CEGZSL}&70.4&63.1&78.6&70.0&77.5&63.9&66.8&65.3&63.3&48.8&38.6&43.1 \\
DFCA-GZSL~\citep{Su2023DualalignedFC}&74.7&66.5&81.5&73.3&\textbf{80.0}&70.9&63.1&66.8&62.6&48.9&38.8&43.3\\

LisGan-CPM \citep{zhang2022zero} & 77.6 & 60.4 & 73.7 & 66.4 & 61.9 & 45.9 & 55.2 & 50.1 & 65.8 & 45.8 & 37.8 & 41.4 \\

\hline
\end{tabular}}
\vspace*{5mm}
 \caption{ We compare our approach to the current state of the art on the AWA2, CUB, and SUN datasets for both GZSL and ZSL settings. We separately analyze CNN-based and transformer-based backbones to provide a head-to-head evaluation of different methods. The results are categorized into four groups: embedding-based, end-to-end embedding-based (separately for CNN and ViT backbones), and generative. Models marked with the $\dagger$ symbol are explicitly based on a NeSy approach. We highlight the \textbf{best} and \underline{second-best} method for each metric, separately for CNN and ViT-based architectures. For FLVN and FLPN, we report the mean, standard deviation and best results (in parenthesis) over three repetitions.
 }
\label{tab:table_test}
\end{table*}

In ~\autoref{tab:table_test}, we experimentally compare the proposed method against previously published techniques on the AWA2, CUB and SUN datasets. We compare against \textit{embedding-based} methods such as PROTO-LTN~\citep{Martone2022PROTOtypicalLT}, DEM~\citep{dem} and VSE~\citep{vse}, recent embedding-based methods trained \textit{end-to-end} such as AMGML~\citep{AMGML}, CC-ZSL~\citep{CC-ZSL}, CoAR-ZSL~\citep{Du2022BoostingZL}, FLVN~\citep{Manigrasso2023FuzzyLV}, PRZSL \citep{yi2024prototype}, AS-ZSL \citep{zhou2023attribute}, HRT \citep{cheng2023hybrid}, CSDR \citep{gao2024improving}, and CTPM \citep{chen2023integrating}, and \textit{generative} methods such as Cycle-CLSWGAN~\citep{cycle-CLSWGAN}, TGMZ~\citep{TGMZ}, CEGZSL~\citep{CEGZSL}, DFCA-GZSL~\citep{Su2023DualalignedFC} and LisGan-CPM \citep{zhang2022zero}. We further differentiate between methods that employ CNN and transformer backbones.The results of state-of-the-art methods are derived from the respective publications. For FLVN and FLPN, we report the mean, standard deviation, and best results over three repetitions.

FLPN obtains better performance compared to our previous NeSy frameworks PROTO-LTN~\citep{Martone2022PROTOtypicalLT} and FLVN~\citep{Manigrasso2023FuzzyLV}, reaching state-of-the-art results on the AWA2 and SUN datasets. Using the standard LTN grounding, FLPN outperforms FLVN~\citep{Manigrasso2023FuzzyLV} with an absolute improvement of +2.5\% and +5.9\% on AWA2 (H=76.61 vs. 74.1) and SUN (H=45.00 vs. 39.1), and a +3.7\% and +5.96\% on unseen classes (AWA2: U=70.79 vs. 67.1; SUN: U=38.86 vs. 32.9). The greatest difference between FLPN and FLVN is observed in SUN, which consists of more complex scene images, whereas CUB and AWA2 are object-centric approaches. A more detailed comparison of the three architectures is provided in Section \ref{ssec:arch_comparison}. Log-FLPN, which is trained in logarithmic space, achieves a similar performance boost over FLPN (AWA2: H=75.08, U=69.32; SUN: H=44.43, U=54.94). However, when comparing FLPN and Log-FLPN, we do not observe a consistent performance benefit associated with operating in the logarithmic space. Lastly, switching from a CNN- to a ViT-based backbone is associated with a substantial absolute performance benefit of roughly +12\% on both AWA2 and SUN, but not on CUB (-0.36\%). These trends are consistent with those observed in~\citep{Du2022BoostingZL}, and may be partly attributed to the smaller size of the dataset compared to AWA2.

Among architectures based on CNN backbones,  FLPN achieves performance that is better or similar to existing state-of-the-art embedding-based models. The closest methods, in terms of performance, include CC-ZSL~\citep{CC-ZSL}, CoAR-ZSL~\citep{Du2022BoostingZL},  PRZSL~\citep{yi2024prototype}, CSDR~\citep{gao2024improving} and AS-ZSL~\citep{zhou2023attribute}. Generally speaking, FLPN achieves the best performance on AWA2 and SUN, but lags behind methods such as CoAR-ZSL~\citep{Du2022BoostingZL} and CSDR~\citep{gao2024improving} on CUB.

Overall, FLPN shows greater performance benefits in combination with CNN than ViT, but still remains competitive among transformer-based architectures.  For the transformer-based category, ZeroMamba~\citep{hou2024zeromamba} ,TransZero++~\citep{chen2022transzero++} and ZSLViT~\citep{ran2023zero} achieve strong ZSL accuracy on AWA2 (T1=80.0 and 90.1, respectively), yet exhibit a more pronounced drop in H for GZSL (74.2 and 57.6, respectively), confirming that FLPN provides a more balanced seen/unseen classification.

When operating in a GZSL setting, it is crucial to balance performance on seen vs. unseen classes. On AWA2, FLPN (U=70.79) outperforms most other methods on unseen classes, including CC-ZSL (U=62.2), PRZSL (U=65.8),  AS-ZSL (U=66.5) and CoAR-ZSL (U= 68.1). While methods such as  CoAR-ZSL (U=68.1) and HRT (U=78.7) achieve comparable or better results on unseen classes, their performance suffers on seen classes compared to FLPN (S=88.21 vs. 79.1 vs. 58.9). In a GZSL setting, relying on a fixed alignments (as in PRZSL) and subspace embeddings (as in AS-ZSL) limit their ability to balance seen and unseen class representations, where FLPN's adaptive prototypes and fine-grained loss provide a distinct advantage.

Although similar general trends are observed on SUN, on the CUB dataset CSDR~\citep{gao2024improving} and CoAR-ZSL~\citep{Du2022BoostingZL} achieve the best performance in both ZSL (CSDR: T1=80.2; CoAR-ZSL: T1=79.2; FLPN: T1=75.01) and GZSL settings (CSDR: H=75.9; CoAR-ZSL: H=74.0; FLPN: H=75.75), despite FLPN's strong performance on unseen classes (CSDR: U=73.1; CoAR-ZSL: U=70.9; FLPN: U=71.65). This discrepancy could be due to the different methods each model uses to extract image attributes. CoAR-ZSL and CDSR employ a contrastive loss, whereas FLPN focuses on classifying attributes via the \texttt{hasAttribute} predicate. This emphasis on classification might limit FLPN's ability to accurately capture image attributes. CoAR-ZSL's contrastive loss enhances feature separation and attribute discrimination in the embedding space, beneficial for fine-grained datasets like CUB. It uses a dual attention pathway to better align global and local features with semantic attributes, surpassing FLPN's single-path attention.

Compared to generative-based methods, FLPN and Log-FLPN outperform Cycle-CLSWGAN~\citep{cycle-CLSWGAN}, LisGan~\citep{LisGAN}, and LisGan-CPM~\citep{zhang2022zero} for unseen categories. TGMZ~\citep{TGMZ} performs well in AWA2 (T1 = 78.4) in the ZSL setting, but drops in the CUB and SUN datasets and the GZSL setting in general. CEGZSL~\citep{CEGZSL} and DFCA-GZSL~\citep{Su2023DualalignedFC} are competitive but less robust than FLPN. Generative approaches can be effective in bridging the gap between seen and unseen classes, as they artificially enrich the dataset. However, generating class-attribute distributions and synthetic data is computationally costly and resource-intensive, whereas FLPN through Nesy integration effectively embeds prior knowledge in the feature representation with minimal overhead.

In terms of computational performance,  FLPN has an inference time of 0.55 s/im, which is higher than FLVN (0.22 s/im) and Proto-LTN (0.09 s/im), yet lower than those obtained by CC-ZSL (0.85 s/im) and CoAR-ZSL (0.98 s/im). These results demonstrate how FLPN manages to embed a rich set of knowledge --- made possible by the additional axioms--- in a compact embedding space, achieving competitive performance both in terms of accuracy and inference time. 

\subsection{Ablation experiments}
The ablation experiments presented in Table~\ref{tab:table_ablation} were performed to elucidate the effect of individual axioms on FLPN performance. Starting from a baseline $\KK$ that included only axioms $\axclass{}$, we progressively expand it until the whole $\KK$ is obtained. In general, all axioms in combination contribute to a boost in performance in unseen classes, with performance increasing from 60.2 to 71.0 with the CNN backbone (+6.7) and from 72.3 to 82.4 with the ViT backbone (+10.1). Each individual axiom includes different predicates, which are grounded by separate components of the architecture, also depending on the backbone used to extract features. Thus, the knowledge base and its predicates are intrinsically linked to specific sections of the FLPN architecture. Thus, ablating an axiom from the knowledge base is equivalent to ablating the parts of the architecture that ground all predicates that are no longer included in knowledge base, and as a consequence in the loss. In Table \ref{tab:table_ablation}, the model column specifies which parts of the network are included in the training of each ablated model.

\begin{table*}[t]
\centering
\resizebox{\textwidth}{!}{%
\begin{tabular}{p{6cm}|cccccc|cccc|cccc}
\hline
&\multicolumn{6}{c}{Knowledge base}&\multicolumn{4}{c}{AWA2}&\multicolumn{4}{c}{CUB} \\
\hline
\raggedright Architecture &\multicolumn{1}{c}{$\axclass{}$}&\multicolumn{1}{c}{$\axmacro{}$}&\multicolumn{1}{c}{$\axmasked{}$}&\multicolumn{1}{c}{$\axattr{}$}&\multicolumn{1}{c}{$\axsimil{}$}&\multicolumn{1}{c}{$\axdissimil{}$}&T1&U&S&H&T1&U&S&H \\
\hline

\raggedright Resnet101 + $\CP$ & \checkmark &-&-&-&-&-&64.3&60.2&82.5&68.8 &72.9 &67.9&72.2 &70.0 \\

\raggedright Resnet101 + $\CP$ &\checkmark &-&\checkmark&-&-&-&66.3	&62.2	&84.7	&71.7 &74.4 &70.0 &81.0 &75.1 \\

\raggedright Resnet101 + $\CP$ + $\AP$ &\checkmark&-&\checkmark&\checkmark&-&-&69.8	&66.4&	80.2	&72.7 &74.3 &70.1 &80.3 &74.9 \\

\raggedright Resnet101 + $\CP$ + $\AP$ + $\MP$&\checkmark&\checkmark&\checkmark&\checkmark&-&-&70.9	&67.7&	79.3&	73.1 &75.2 &71.6 &80.6 &75.8 \\

\raggedright Resnet101 + $\CP$ + $\AP$ + $\MP$  &\checkmark&\checkmark&\checkmark&\checkmark&\checkmark&\checkmark&72.9&71.0&78.4&74.5 &75.2 &71.7 &78.9 &75.1 \\

\hline

\raggedright ViT + $\CP$ &\checkmark &-&-&-&-&-&72.6 &72.3 &86.2 &78.6 &72.7 &65.3 &76.2 &70.3 \\

\raggedright ViT + $\CP$ &\checkmark &-&\checkmark&-&-&-&79.8 &78.9 &89.6 &83.9 &73.3 &66.5 &78.3 &71.9 \\

\raggedright ViT + $\CP$ + $\AP$ &\checkmark&-&\checkmark&\checkmark&-&-&80.9 &80.3 &89.2 &84.5 &76.2 &72.4&73.5 &72.9 \\

\raggedright ViT + $\CP$ + $\AP$ + $\MP$ &\checkmark&\checkmark&\checkmark&\checkmark&-&-&83.1 &82.6 &85.2 &83.9 &77.5 &71.4 &74.3 &72.8 \\

\raggedright ViT + $\CP$ + $\AP$ + $\MP$ &\checkmark&\checkmark&\checkmark&\checkmark&\checkmark&\checkmark&83.0 &82.4 &87.6 &84.9 &77.6 &73.7 &75.3 &72.1 \\

\hline
\end{tabular}}
\vspace*{5mm}
\caption{Ablation study: effect of different axioms in the  $\KK$ on the performance, for both CNN (top rows) and ViT (bottom rows) backbones, on Awa2 and CUB.}
\label{tab:table_ablation}

\end{table*}

\textbf{Role of $\axclass{}$}. The base architecture includes only axioms designed to learn from labeled examples. Each class has inherent characteristics that are always present: for instance, a zebra will always have stripes, even in a cropped image. As a result, the network prioritizes seen classes. 

\textbf{Role of $\axmasked{}$}. This axiom accounts for exceptions by randomly masking attributes ($\axmasked{}$).
It improves on both seen and unseen classes (S=84.7 for AWA2 and S=81.0 for CUB in Table~\ref{tab:table_ablation}).
The introduction of axiom $\axmasked{}$ helps the model to handle exceptions due to the use of image-level attribute labels during  training.  In preliminary experiments, we found that $\axmasked{}$ is particularly effective when the $k$ elements are chosen among attributes with values lower than the class average. 

\textbf{Role of $\axattr{}$}. Using the attribute matrix and the attribute weight $w_{a}$ for a class, we introduce the attribute classification with the axiom $\axattr{}$. The predicate $\texttt{hasAttribute}$ in this axiom is based on the cosine distance between an attribute-level embedding and an attribute prototype.
Despite a nearly 1-point loss in the GZSL setting, Table~\ref{tab:table_ablation} shows improvement in unseen class classification in the ZSL (T1 = 66.3 to 69.8 for AWA2) and GZSL (U = 62.2 to 66.4 for AWA2 and U = 70.1 to 70.1 for CUB) settings, respectively.

\textbf{Role of $\axmacro{}$}. The inclusion of axiom $\axmacro{}$ slightly improves the harmonic mean (H) from 72.7 to 73.1 for AWA2 and 74.9 to 75.8 for CUB, indicating that the addition of semantically relevant constraints on macroclasses aggregation provides a minor boost to the architecture's performance.Proto-LTN~\citep{Martone2022PROTOtypicalLT} has previously shown that pre-trained embeddings, even without the imposition of specific constraints on the class hierarchy, naturally tend to cluster related classes in feature space.

While previous observations were based on t-Distributed Stochastic Neighbor (t-SNE) visualization, the present ablation study paints a slightly more nuanced picture: explicitly introducing constraints on macroclasses negatively affects performance on seen classes S ($-0.9\%$ with CNN backbone, $-4\%$ with ViT backbone on the AWA2 dataset) but promotes generalization to unseen classes U ($+1.3\%$ with CNN backbone, $+2.3\%$ with ViT backbone also on AWA2). On CUB we observe a slight increase for seen classes with both CNN ($+0.3\%$) and ViT ($+0.8\%$), while generalization on unseen classes improves with CNN ($+1.5\%$) but slightly decreases with ViT ($-1.0\%$). This phenomenon can be attributed to the aggregation of related classes in a more compact feature space, which benefits unseen samples.

\textbf{Role of $\axsimil{}$ and $\axdissimil{}$.} These two axioms introduce constraints that force similar images (i.e., images that are close in the embedding space) to be classified as belonging to the same class, while dissimilar images (i.e., those farther away in the feature space) must be classified into different classes. Both axioms act essentially as a contrastive component to the loss. From Table~\ref{tab:table_ablation}, we observe that these axioms in combination further boost the harmonic mean (H) on AWA2 (+1.4\% with the CNN backbone, +1\% with the ViT backbone), with a performance increase attributed mostly to unseen classes for the CNN backbone (+3.3\%) and to seen classes for the ViT backbone (+2.4\%).
On the CUB dataset, these axioms achieve a more limited performance gain. We hypothesize that this may be due to the fine-grained nature of the CUB classification task.

\begin{figure*}[tb!]
 \centering
 \begin{subfigure}[b]{0.22\textwidth}
 \centering
 \includegraphics[width=\textwidth]{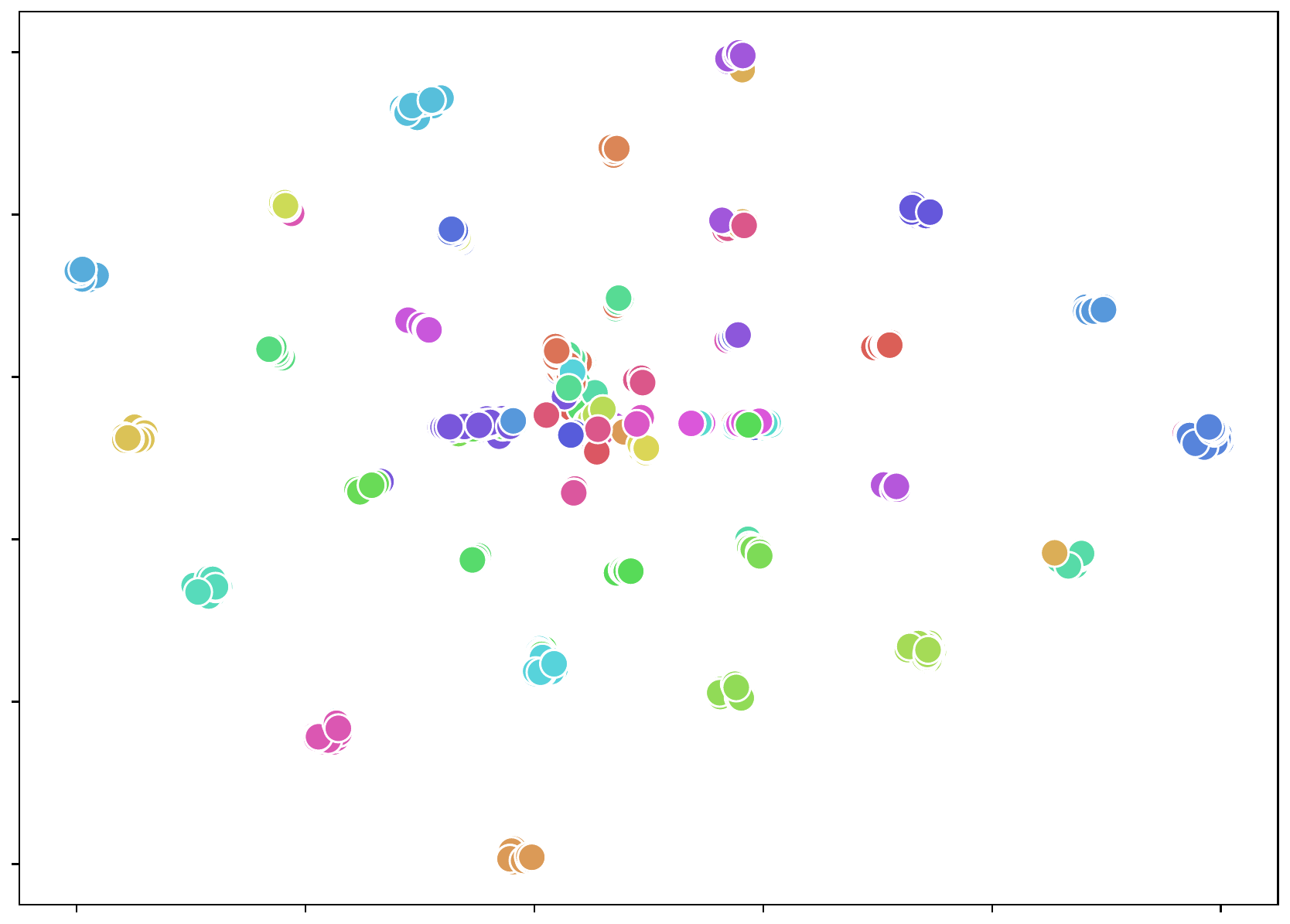}
 \caption{ResNet-101, $\KK_r$, seen classes}
 \label{fig:fig1}
 \end{subfigure}
 \hfill
 \begin{subfigure}[b]{0.22\textwidth}
 \centering
 \includegraphics[width=\textwidth]{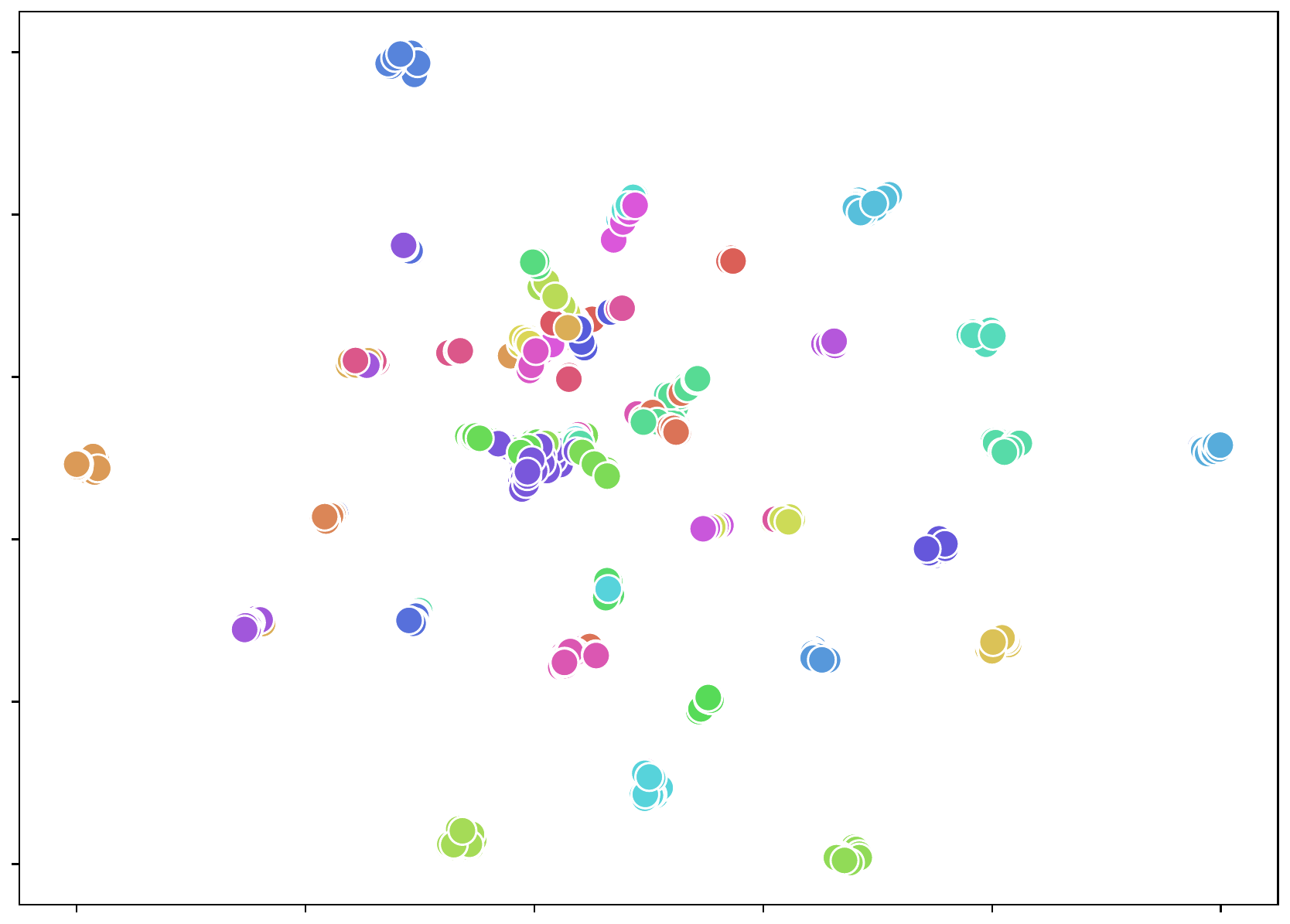}
 \caption{ResNet-101, $\KK$, seen classes}
 \label{fig:fig2}
 \end{subfigure}
 \hfill
 \begin{subfigure}[b]{0.22\textwidth}
 \centering
 \includegraphics[width=\textwidth]{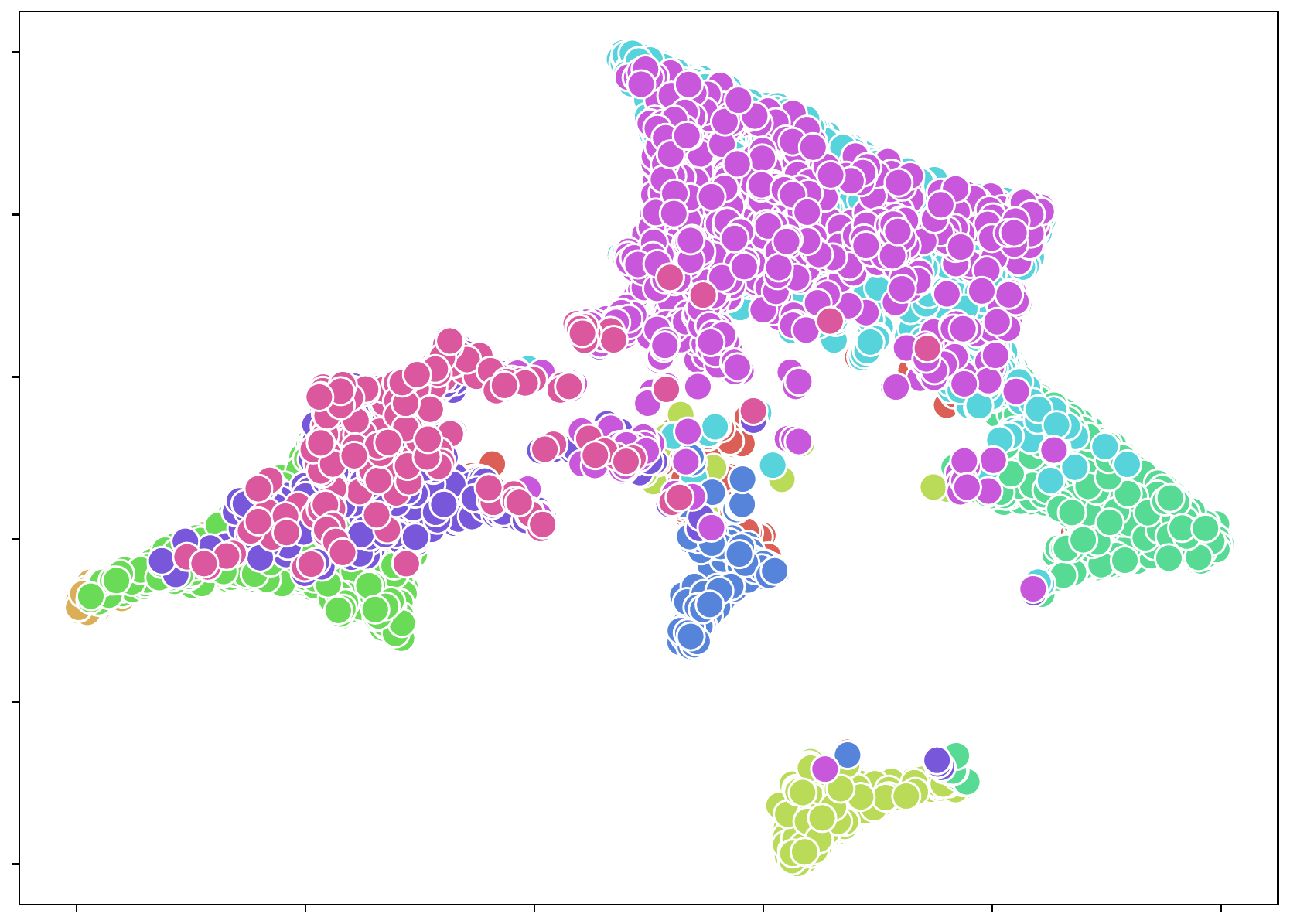}

 \caption{ResNet-101, $\KK_r$, unseen classes}
 \label{fig:fig3}
 \end{subfigure}
 \hfill
 \begin{subfigure}[b]{0.22\textwidth}
 \centering
 \includegraphics[width=\textwidth]{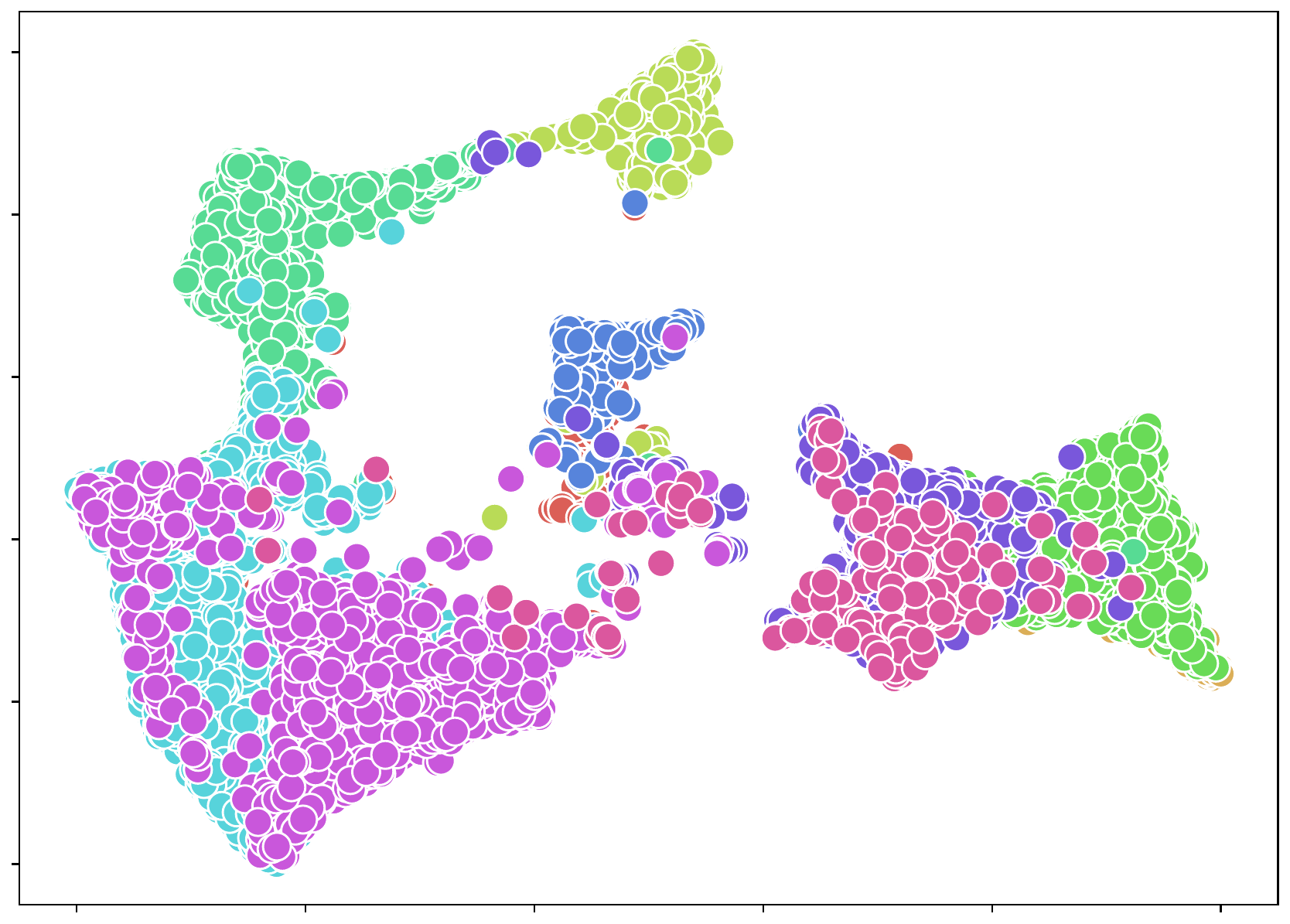}
  \caption{ResNet-101, $\KK$, unseen classes}
 \label{fig:fig4}
 \end{subfigure}
 \\
 \vspace{0.5cm}
 \begin{subfigure}[b]{0.22\textwidth}
 \centering
 \includegraphics[width=\textwidth]{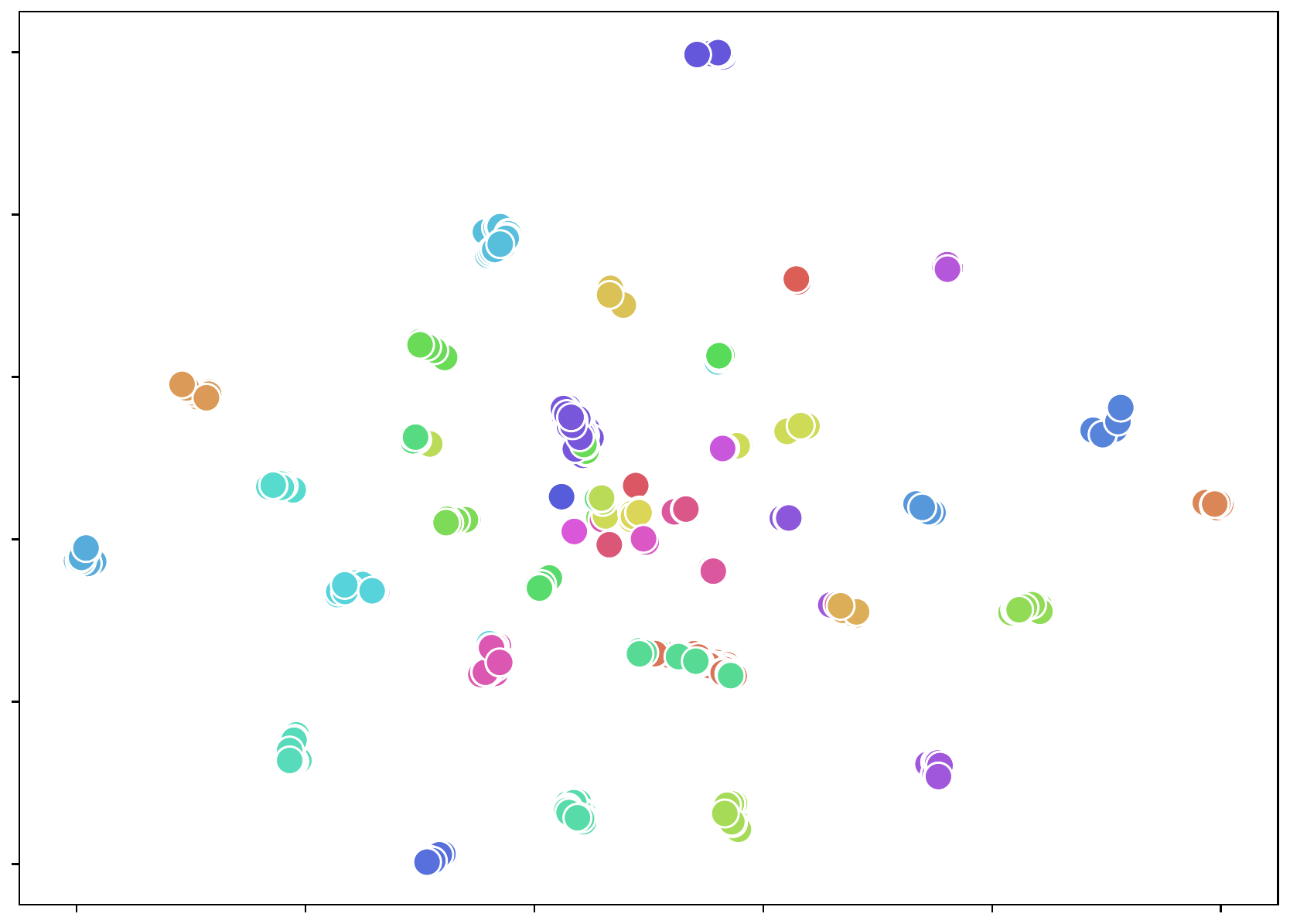}

 \caption{ViT, $\KK_r$, seen classes}
 \label{fig:fig5}
 \end{subfigure}
 \hfill
 \begin{subfigure}[b]{0.22\textwidth}
 \centering
 \includegraphics[width=\textwidth]{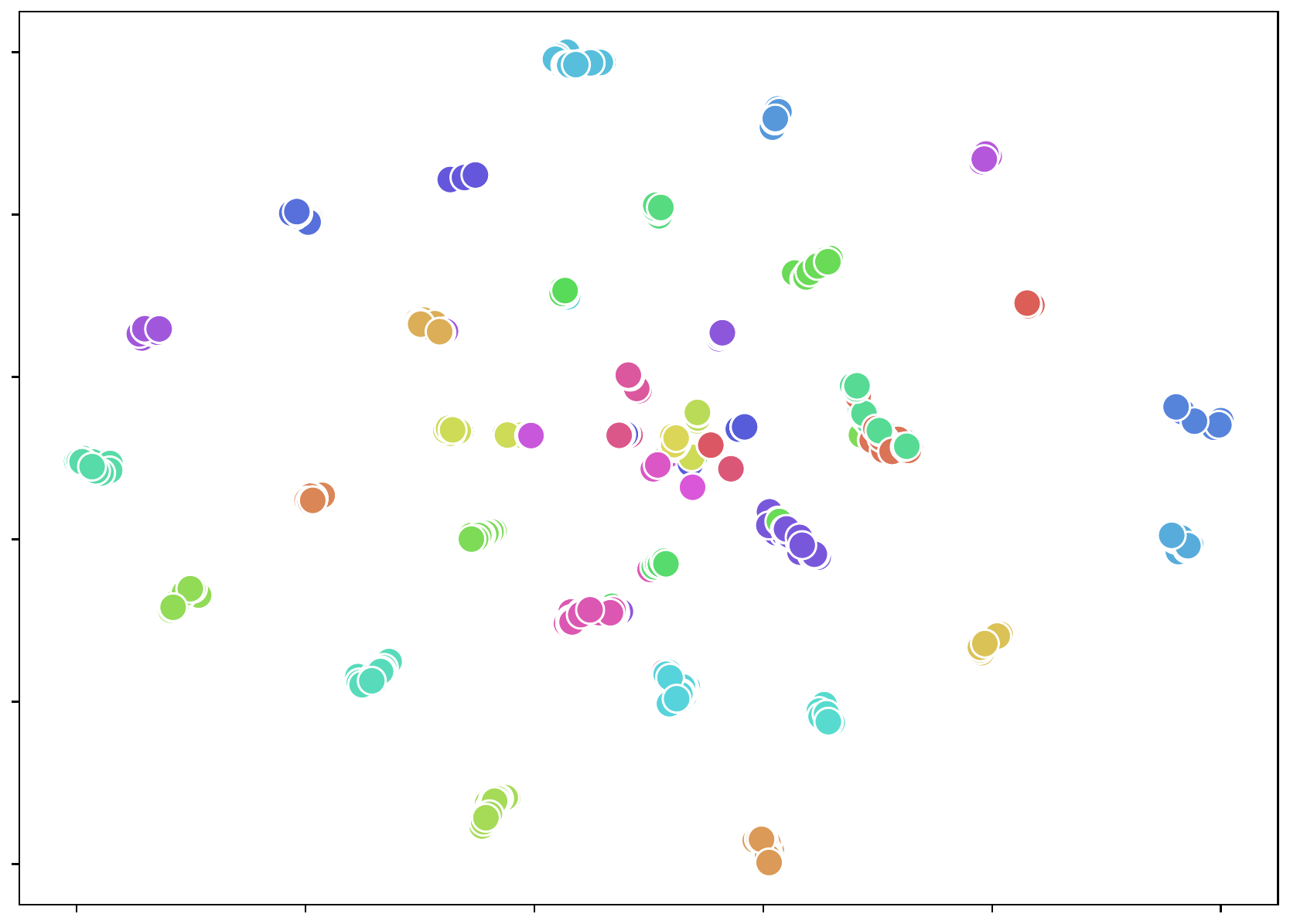}
  \caption{ViT,$ \KK $, seen classes}

 \label{fig:fig6}
 \end{subfigure}
 \hfill
 \begin{subfigure}[b]{0.22\textwidth}
 \centering
 \includegraphics[width=\textwidth]{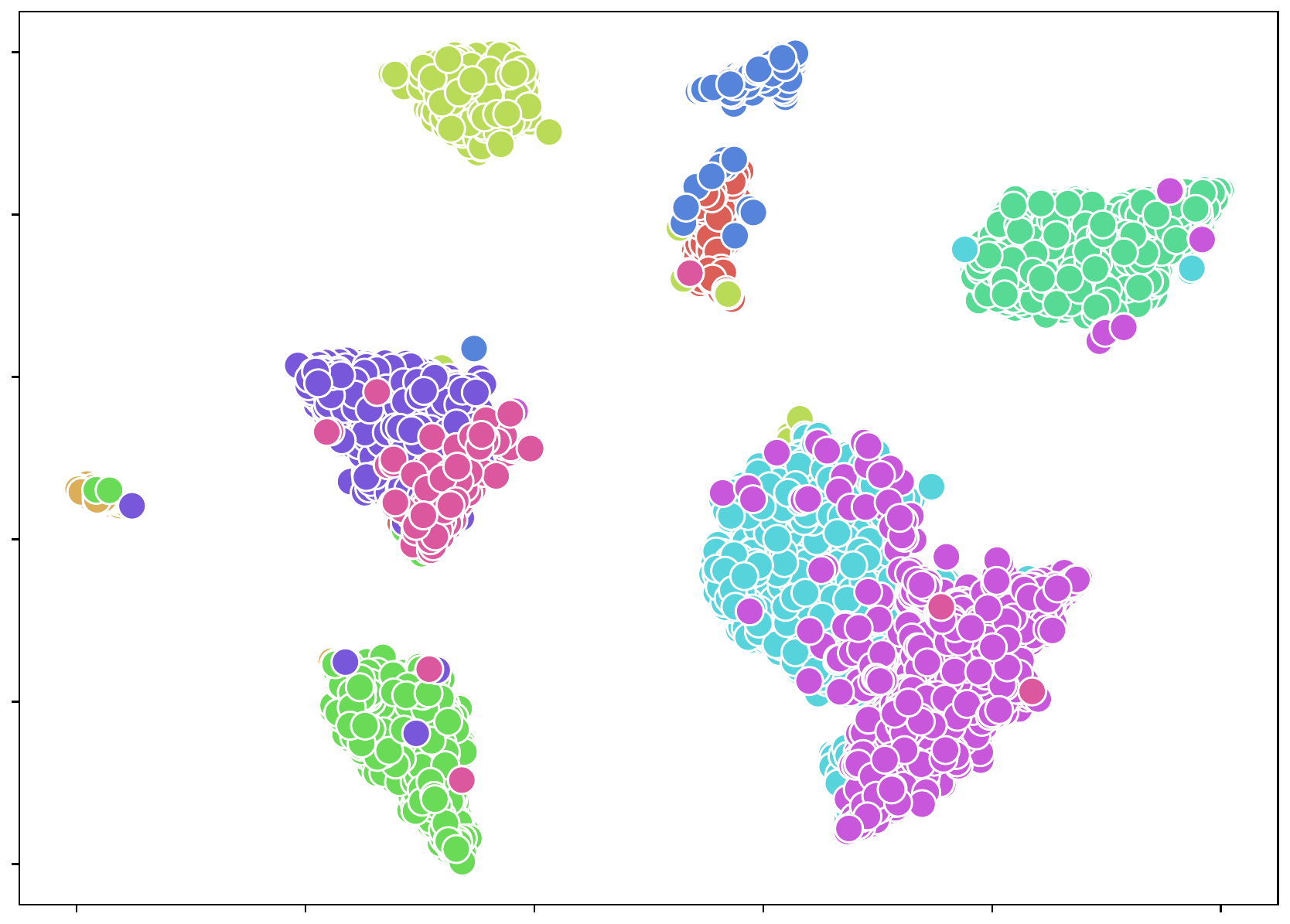}
  \caption{ViT, $\KK_r$, unseen classes}

 \label{fig:fig7}
 \end{subfigure}
 \hfill
 \begin{subfigure}[b]{0.22\textwidth}
 \centering
 \includegraphics[width=\textwidth]{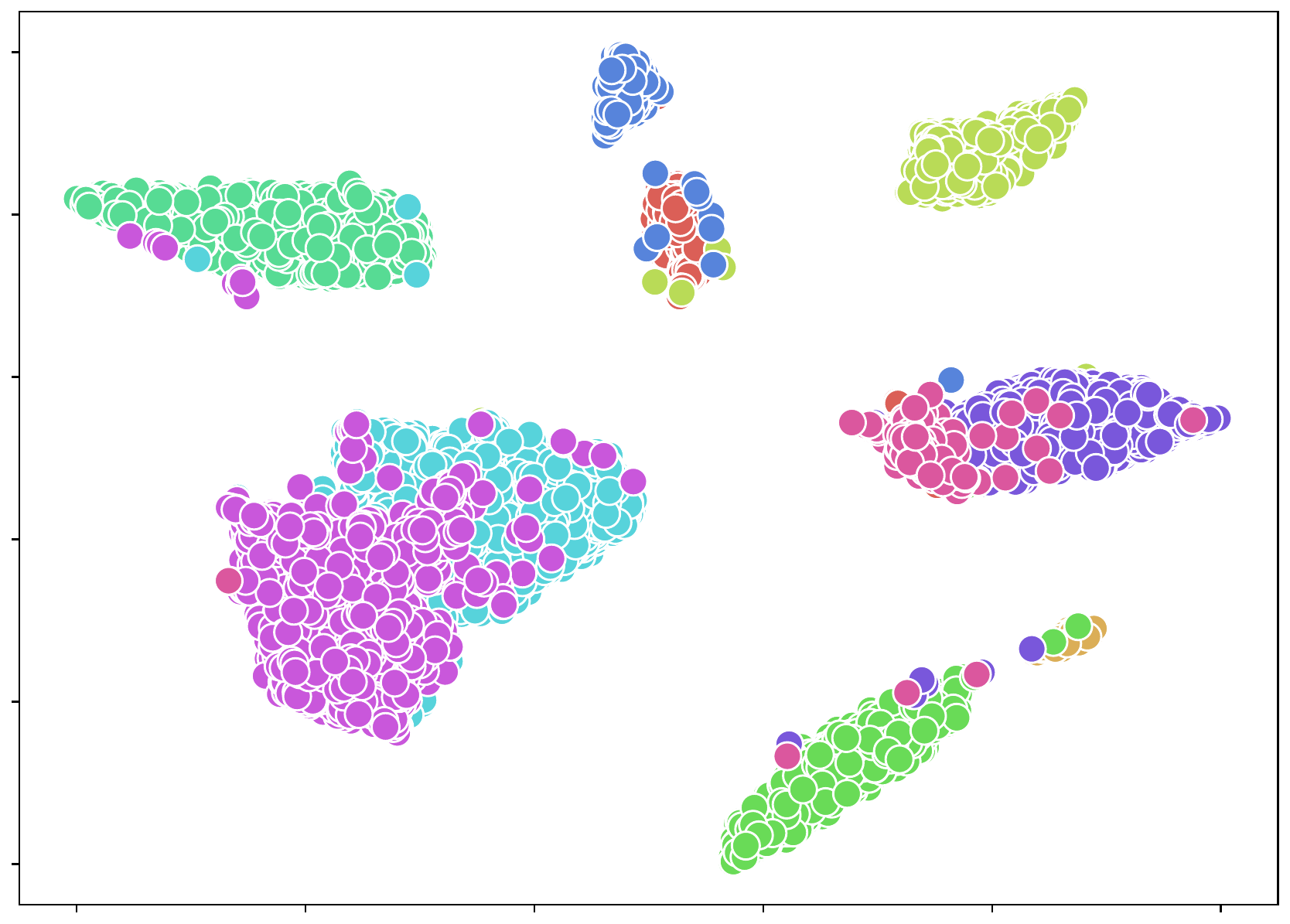}
   \caption{ViT, $\KK$, unseen classes}

 \label{fig:fig8}
 \end{subfigure}
 \caption{The class-level feature distribution on AWA2 dataset for FLPN with the ResNet-101 (a-d) and ViT (e-h) backbones is represented using t-SNE. The models in plots (a), (c), (e), and (g) contain a knowledge base $\KK_r$ composed only of $\axclass{}$ and $\axmasked{}$, whereas the knowledge base in the remaining plots includes all the axioms. The likelihood scores generated by the model for every class were utilized to produce these representations.}
 \label{fig:tsnet_class}
\end{figure*}

\begin{figure*}[tb!]
 \centering

 \begin{subfigure}[b]{0.22\textwidth}
 \centering
 \includegraphics[width=\textwidth]{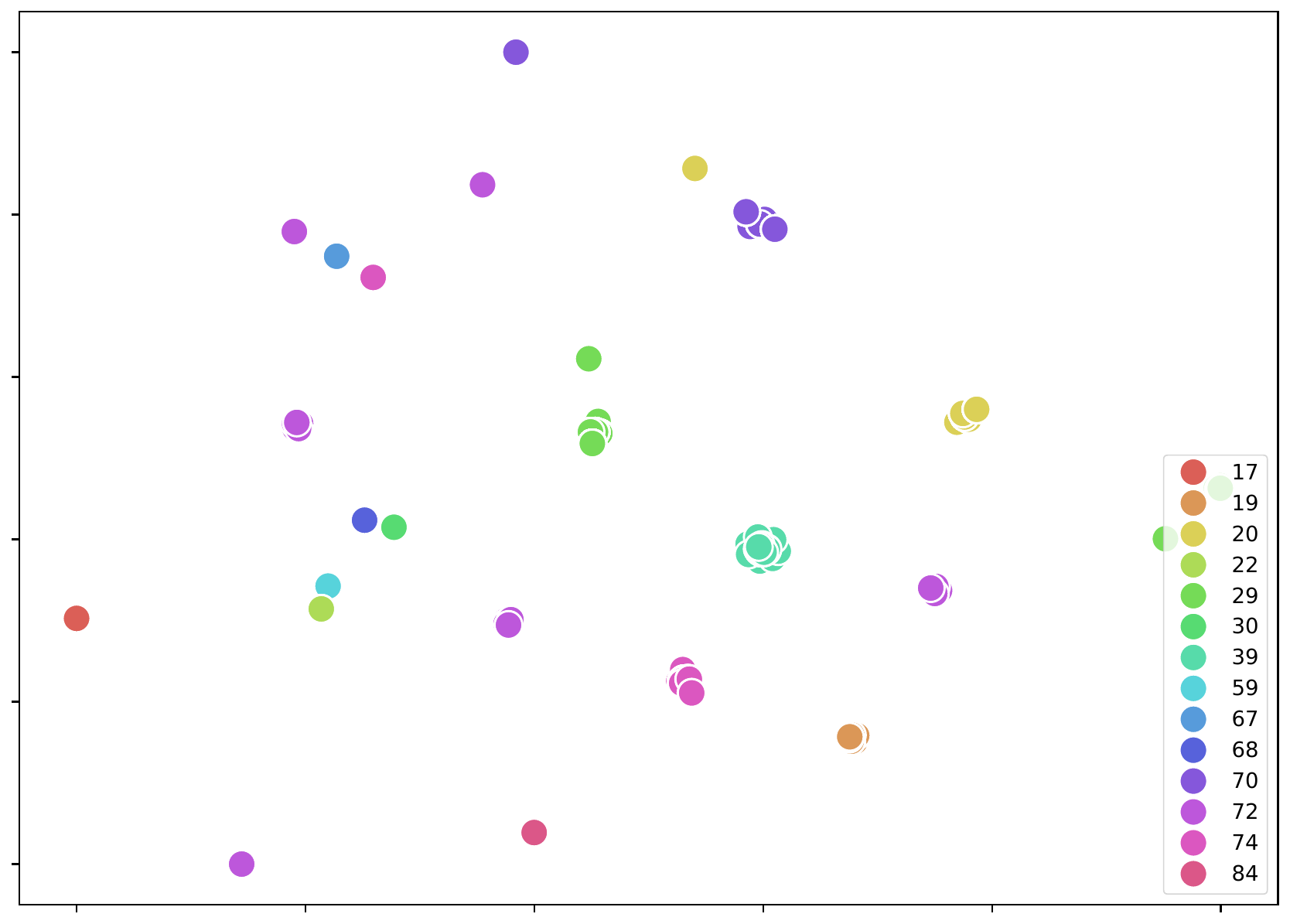}
 \caption{seen classes;  ResNet-101 backbone}
 \label{fig:fig1_tsnet_awa2}
 \end{subfigure}
 \hfill
 \begin{subfigure}[b]{0.22\textwidth}
 \centering
 \includegraphics[width=\textwidth]{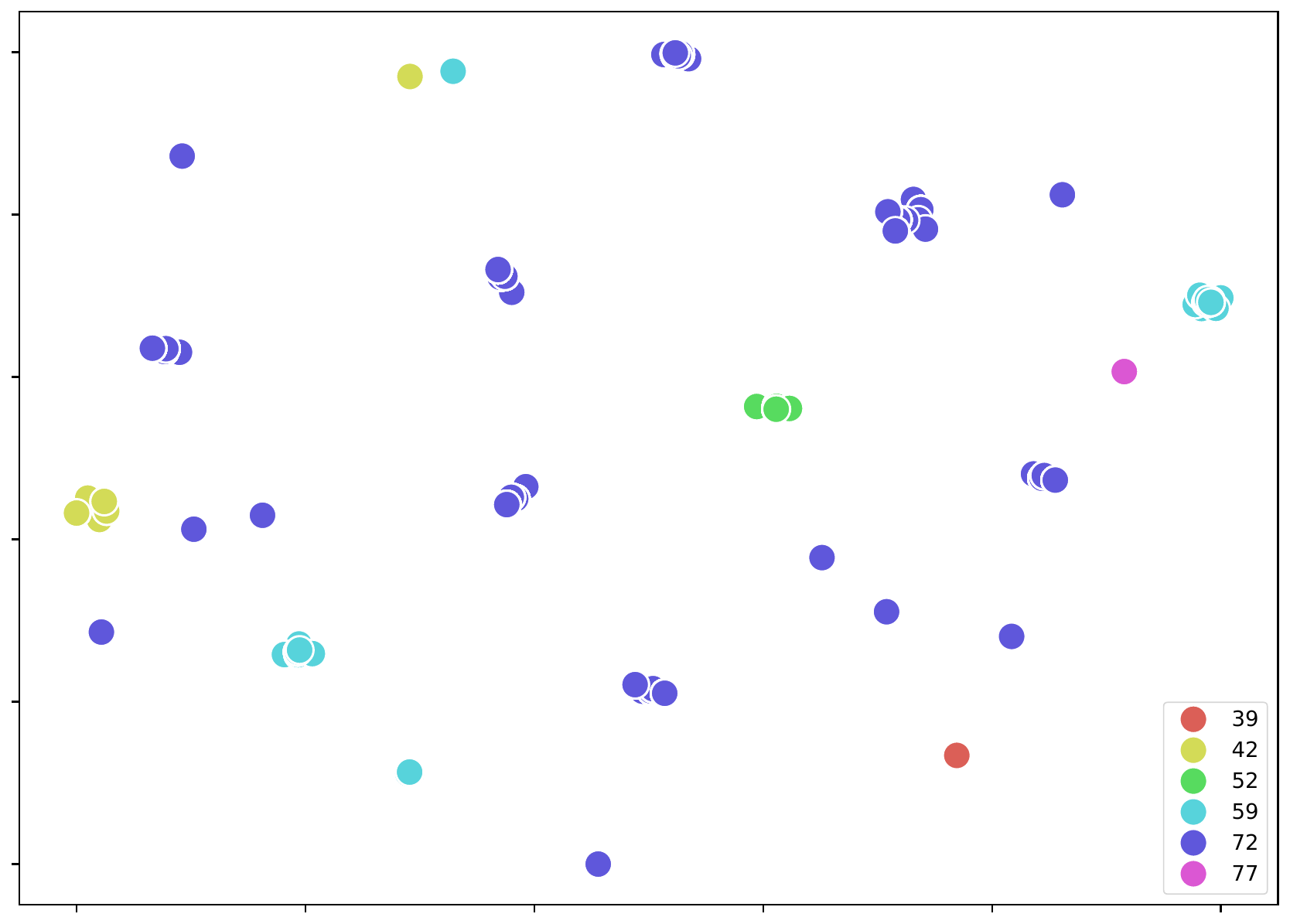}
 \caption{unseen classes;  ResNet-101 backbone}
 \label{fig:fig2_tsnet_awa2}
 \end{subfigure}
 \hfill
 \begin{subfigure}[b]{0.22\textwidth}
 \centering
 \includegraphics[width=\textwidth]{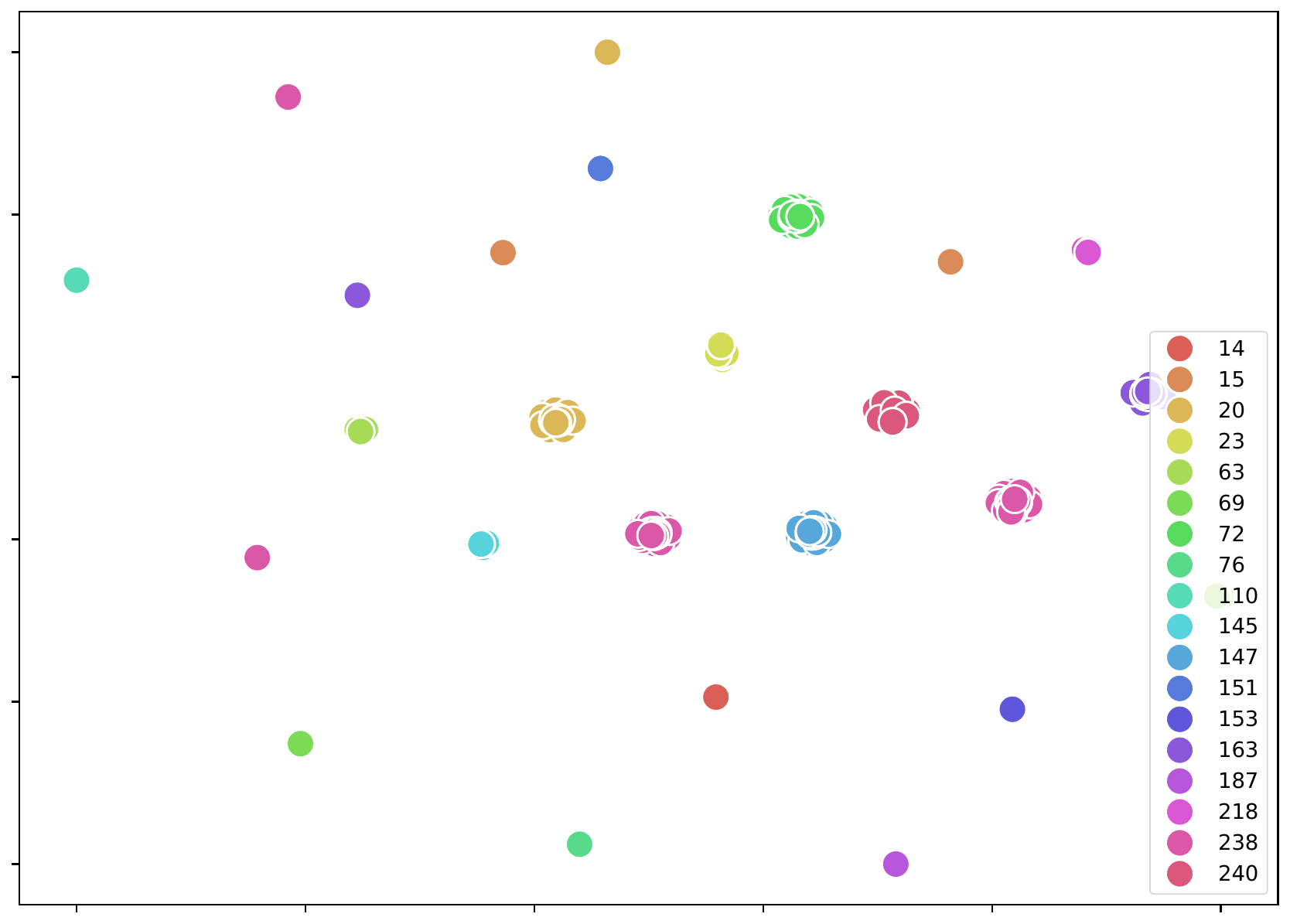}
 \caption{seen classes;  ResNet-101 backbone}
 \label{fig:fig1_tsnet_cub}
 \end{subfigure}
 \hfill
 \begin{subfigure}[b]{0.22\textwidth}
 \centering
 \includegraphics[width=\textwidth]{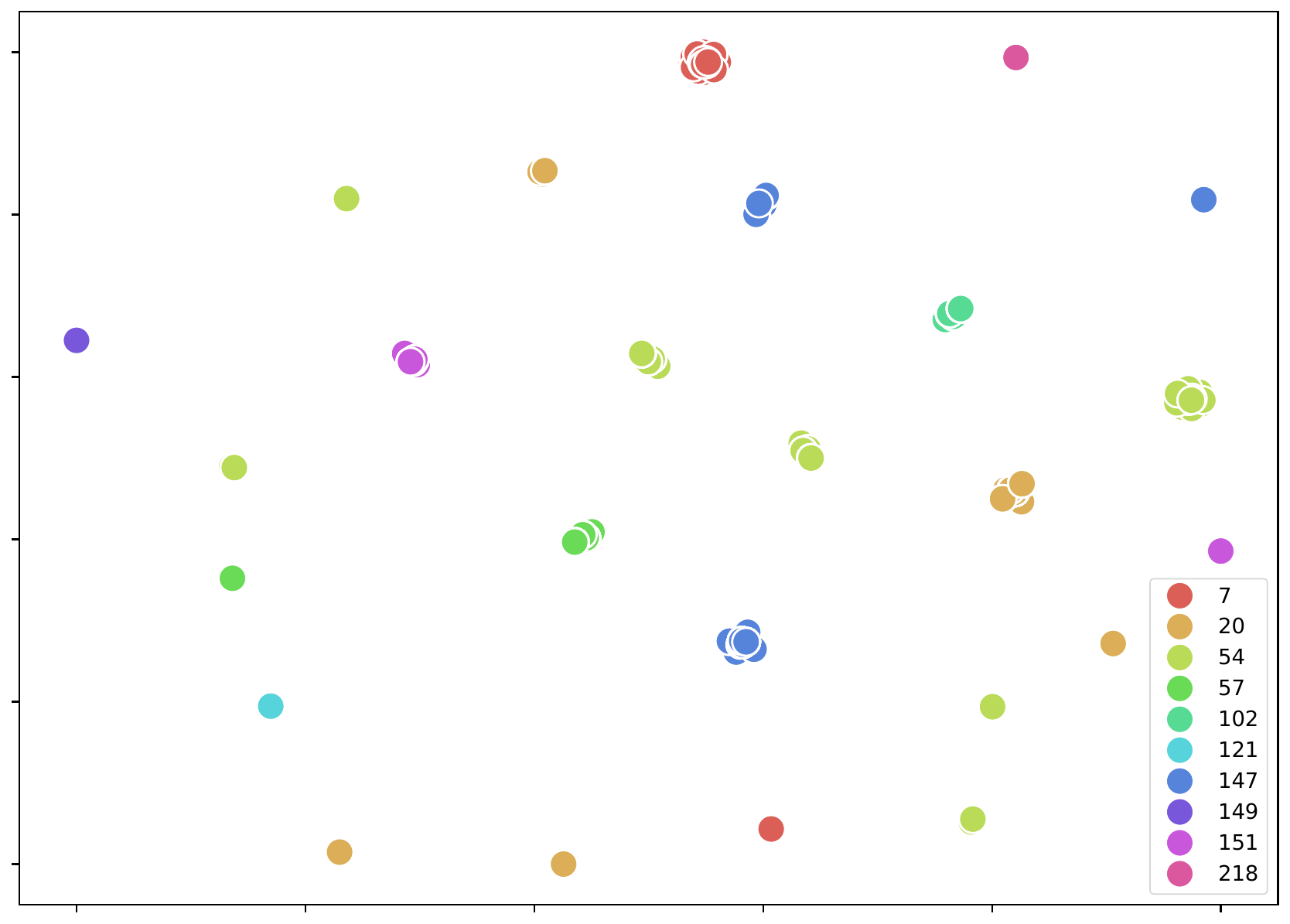}
 \caption{unseen classes;  ResNet-101 backbone}
 \label{fig:fig2_tsnet_cub}
 \end{subfigure}

 \caption{Attribute-level feature distribution for FLPN on AWA2 (a-b) and CUB (c-d) datasets  represented using t-SNE.}
 \label{fig:tsnet_attribute}
\end{figure*}

\subsection{Class feature visualization using T-SNE}

We examined whether the extracted attributes accurately represent the characteristics of the dataset using t-SNE visualization. We visualized class separation on the AWA2 and CUB datasets using attribute scores, as shown in Figure~\ref{fig:tsnet_class}. Focusing on the GZSL scenario, we compared the embeddings for two knowledge bases: one with a minimal set of axioms ($\axclass{}$ and $\axmasked{}$), and the final one with all proposed axioms. We analyzed the embedding space for both seen and unseen classes separately. For clarity, we included only 500 samples per class, each with a distinct color. Although Figure~\ref{fig:fig2} and Figure~\ref{fig:fig6} show a slight improvement in the distribution of unseen classes compared to Figure~\ref{fig:fig3} and Figure~\ref{fig:fig4} (supporting the results in Table~\ref{tab:table_test}), both the seen and unseen classes are clearly separated in the feature space.

Figure~\ref{fig:tsnet_attribute} presents the t-SNE visualizations of the attribute-level embeddings for each image. We analyze both the AWA2 (Figure~\ref{fig:fig1_tsnet_awa2}, Figure~\ref{fig:fig2_tsnet_awa2}) and the CUB datasets (Figure~\ref{fig:fig1_tsnet_cub}, Figure~\ref{fig:fig2_tsnet_cub}), focusing on attributes of seen and unseen classes.
The attention mechanism helps the model prioritize relevant attributes for distinguishing classes. Key attributes, especially in unseen class images, are distinctly grouped in the feature space. This result suggests that the extracted information effectively captures valuable characteristics for accurate class discrimination.

\subsection{Attribute attention maps}
\begin{figure*}[tb!]
 \centering
 \begin{subfigure}{0.15\linewidth}
 \includegraphics[width=\linewidth]{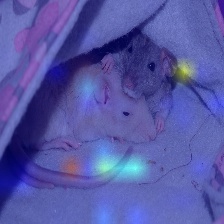}
 \caption{has tail}
 \end{subfigure}
 \begin{subfigure}{0.15\linewidth}
 \includegraphics[width=\linewidth]{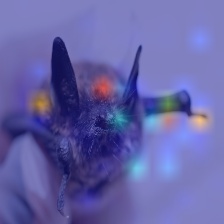}
 \caption{can fly}
 \end{subfigure}
 \begin{subfigure}{0.15\linewidth}
 \includegraphics[width=\linewidth]{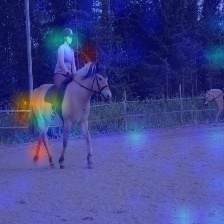}
 \caption{has long leg}
 \end{subfigure}
 \begin{subfigure}{0.15\linewidth}
 \includegraphics[width=\linewidth]{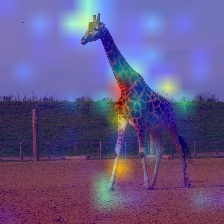}
 \caption{long neck}
 \end{subfigure}
 \begin{subfigure}{0.15\linewidth}
 \includegraphics[width=\linewidth]{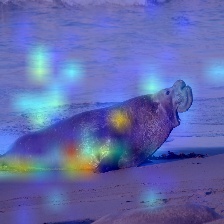}
 \caption{has flippers}
 \end{subfigure}
 \begin{subfigure}{0.15\linewidth}
 \includegraphics[width=\linewidth]{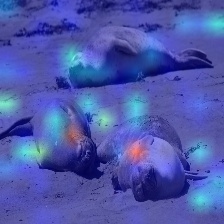}
 \caption{has flippers}
 \end{subfigure}

 \begin{subfigure}{0.15\linewidth}
 \includegraphics[width=\linewidth]{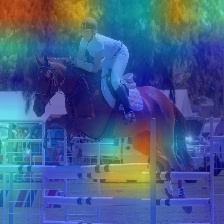}
 \caption{quadruped}
 \end{subfigure}
 \begin{subfigure}{0.15\linewidth}
 \includegraphics[width=\linewidth]{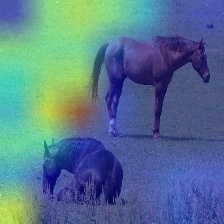}
 \caption{has tail}
 \end{subfigure}
 \begin{subfigure}{0.15\linewidth}
 \includegraphics[width=\linewidth]{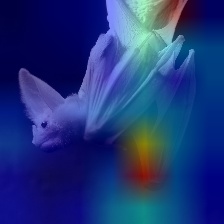}
 \caption{can fly}
 \end{subfigure}
 \begin{subfigure}{0.15\linewidth}
 \includegraphics[width=\linewidth]{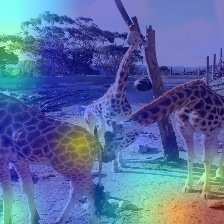}
 \caption{has tail}
 \end{subfigure}
 \begin{subfigure}{0.15\linewidth}
 \includegraphics[width=\linewidth]{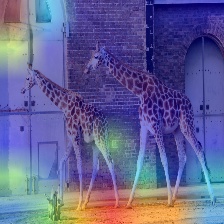}
 \caption{has long leg}
 \end{subfigure}
 \begin{subfigure}{0.15\linewidth}
 \includegraphics[width=\linewidth]{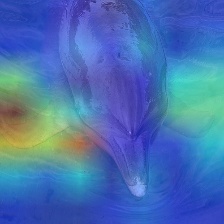}
 \caption{has flippers}
 \end{subfigure}

 \caption{Attention maps for randomly selected examples and attributes from AWA2 with ViT (a-f) and ResNet101 (g-l) backbones.}
 \label{fig:attention_maps}
\end{figure*}

Introducing axioms based on the $\texttt{hasAttribute}$ predicate, which enable FLPN to produce attribute-level classification, demonstrably improves overall image classification results (as shown in Table~\ref{tab:table_test} by comparing against FLVN architecture). This section explores how the attention mechanism helps to identify specific attributes in images for both CNN- and ViT-based architectures (Figure~\ref{fig:attention_maps}). Both the examples showcase the model's ability to pinpoint image regions with specific properties. Qualitatively, the ViT-based architecture seems to outperform the CNN-based one. In both the cases, limitations persist due to the lack of annotations at the attribute level, which, as also noted in prior works~\citep{Du2022BoostingZL}, can introduce noise into the attention maps.

\subsection{Prototypical LTNs family: architectural comparison}
\label{ssec:arch_comparison}

The proposed architecture is an evolution and improvement of the previous architectures Proto-LTN~\citep{Martone2022PROTOtypicalLT} and FLVN~\citep{Manigrasso2023FuzzyLV}, both based on similar NeSy principles. Table~\ref{tab:table_differences2} details differences among the three architectures concerning the knowledge base $\KK$ used for training the models, the \textit{embedding space}, that is the shared embedding space used to compute the similarity between the image embeddings and the class prototypes, the \textit{grounding} of the $\texttt{isOfClass}$ predicate, how the \textit{Batch} is constructed by selecting examples for the training set, the \textit{Aggregator} which combines all axioms to compute the final satisfiability, and lastly the \textit{backbone}.

\textbf{Proto-LTN vs FLPN} Proto-LTN~\citep{Martone2022PROTOtypicalLT} differs both in the knowledge base and in the implementation of the $\texttt{isOfClass}$ predicate. Proto-LTN leverages the image space as a shared embedding, with training involving random sample selection. It employs the Euclidean distance to reinforce similarity between images and prototypes belonging to the same class. In contrast, FLPN adopts a  similarity metric that multiplies image and prototype features, followed by a Softmax activation function. This formulation, incorporating mutual exclusivity constraints, facilitates end-to-end training without features collapsing or overfitting to the seen classes.
Unlike Euclidean distance-based measures, classifying an image into one class decreases the likelihood that the same embedding will be assigned to other classes.

\textbf{FLVN vs FLPN} In addition to their shared embedding space and backbone architecture, FLVN and FLPN exhibit fundamental design similarities. Some constraints inherent to FLVN structure preclude the implementation of a prototypical network.  While FLVN incorporates attribute information by multiplying image features (projected into the attribute space) with the semantic attribute matrix, it does not explicitly generate class prototypes. As a result, directly incorporating a prototypical network for attributes within the FLVN architecture may not be feasible. In contrast, FLPN includes multiple prototypical networks, which however share initial layers to reduce the effective parameter count and prevent overfitting.

\begin{table*}[tb!]
\centering
\resizebox{\textwidth}{!}{%
\begin{tabular}{c|cccccc}
\hline
Model & Knowledge base & Embedding & End-to-end & $\texttt{isOfClass}$ grounding & Batch sampling & Aggregator \\
 & & space & training & & & \\
\hline
Proto-LTN & \{$\axclass{}$\} & $M$ & - & Euclidean & Random & LogProduct \\
FLVN & \{$\axclass{},\axmacro{},\axmasked{},\axsimil{},\axdissimil{}$\} & $A$ & \checkmark & Softmax & Shots and Ways & Mean-Error \\

FLPN (ours) & \{$\axclass{},\axmacro{},\axmasked{},\axattr{},\axsimil{},\axdissimil{}$\} & $M$ & \checkmark & Softmax & Random & Mean-Error \\
\hline
\end{tabular}}
\vspace*{5mm}
\caption{Comparison of FLPN, FLVN and Proto-LTN NeSy architectures based on Res101 backbone}
\label{tab:table_differences2}
\end{table*}

\section{Conclusion}
\label{sec:conclusion}
Building on prior works in ZSL~\citep{CC-ZSL,Du2022BoostingZL} and embedding them within a NeSy framework ~\citep{Martone2022PROTOtypicalLT,Manigrasso2023FuzzyLV}, we developed a new NeSy architecture for ZSL and GZSL tasks named FLPN. Like FLVN, FLPN integrates axioms combining class-specific knowledge (e.g., class hierarchies) with inductive biases to handle dataset exceptions (e.g., ``a non-ferocious lion'') and establish image relationships (e.g., similar images belong to the same class). These axioms serve as semantic priors, addressing annotation gaps by considering class-level attributes and weighting image-level attributes, strengthening the NeSy foundation for GZSL tasks.
Experimental results show that FLPN meets or exceeds existing methods on standard GZSL benchmarks with respect to both performance and explainability. FLPN does not require multiple backbone networks and maintains a parameter count comparable to traditional embedding techniques. The proposed approach can be integrated into other techniques, e.g., by modifying the predicate grounding accordingly.  We see several expansion avenues: exploring different formulations for the $\texttt{isOfClass}$ predicate (e.g., based on hyperbolic embeddings~\citep{khrulkov2020hyperbolic}), incorporating alternative attribute detection mechanisms, introducing contrastive loss as a logical constraint, expanding the knowledge base to include class-related information beyond class hierarchies, incorporating information from sources such as language models, or testing FLPN in a transductive ZSL context.

\section*{Acknowledgments}
This study was carried out within the FAIR - Future Artificial Intelligence Research and received funding from the European Union Next-GenerationEU (PIANO NAZIONALE DI RIPRESA E RESILIENZA (PNRR) -- MISSIONE 4 COMPONENTE 2, INVESTIMENTO 1.3 -- D.D. 1555 11/10/2022, PE00000013). This manuscript reflects only the authors' views and opinions, neither the European Union nor the European Commission can be considered responsible for them.

\bibliographystyle{elsarticle-num}
\bibliography{main}

\end{document}